\documentclass[runningheads]{llncs}

\usepackage{eccv}

\usepackage{eccvabbrv}

\usepackage{graphicx}
\usepackage{booktabs}
\usepackage{pifont} 
\usepackage[accsupp]{axessibility}  

\usepackage{hyperref}

\usepackage{orcidlink}
\usepackage[table]{xcolor}

\begin{document}

\title{AVBench: Human-Aligned and Automated Evaluation Benchmark for Audio-Video Generative Models} 

\titlerunning{AVBench}

\author{Jialiang Yang\inst{1} \and
Bin Xia\inst{2} \and
 Ruihang Chu\inst{1}\and
  Dingdong Wang\inst{2}\and
  Wanke Xia\inst{1}\and
  Zhun Mou\inst{1}\and
Tianyang Zhong\inst{1}\and
  Yiting Zhao\inst{1}\and
 Wenming Yang\inst{1}\thanks{Corresponding author}}


\authorrunning{F.~Author et al.}

\institute{Tsinghua University \and
The Chinese University of Hong Kong\\
\vspace{2pt} 
\small Project Page: \url{https://yajialiang.github.io/AVBench-site/}}


\maketitle

\begin{abstract}

Rapid advances in audio-video (AV) generation have enabled high-fidelity synthesis with synchronized sound, particularly for human-related scenarios involving speech and interactions.
Yet evaluation for AV generation remains at an early stage, with only a few coarse-grained benchmarks for human-related scenarios and relying on limited preset evaluations with generic multimodal LLMs, leading to inaccurate assessments of model capabilities.
To address these issues, we introduce AVBench, a fully automated benchmark tailored for human-centric AV generation. AVBench is built on two key designs for comprehensive and accurate evaluation: (i) \textbf{Human-centric and fine-grained metrics.} AVBench integrates ten evaluation dimensions designed for human-centered real-world scenarios, covering visual quality, audio quality, and multi-level consistency across modalities. These practical metrics capture human-related details that existing benchmarks often overlook.
(ii) \textbf{Specialized evaluators via preference learning.} To address the lack of specialized training data, we construct large-scale supervision by transforming real-world videos into diverse training pairs with controlled perturbations. After fine-tuning on this high-quality dataset, the evaluators learn to reliably detect subtle cross-modal inconsistencies.
Crucially, instead of producing discrete textual judgment, AVBench derives continuous evaluation scores from the model’s prediction confidence on binary decisions. This probabilistic scoring mechanism enables a more reliable assessment than traditional VQA-style evaluation and aligns closely with human judgment.
Taken together, AVBench offers automated evaluation for AV generation, demonstrates strong potential for data filtering and serving as a differentiable reward signal for Reinforcement Learning from Human Feedback (RLHF).
\end{abstract}

\section{Introduction}
Text-to-Audio-Video (T2AV) generation offers more natural audio results compared to previous Text-to-Video (T2V) generation, providing users with a better experience and achieving widespread success, such as Veo3~\cite{wiedemer2025veo3}, and has become a new trend in industrial development. State-of-the-art (SOTA) systems like Sora2~\cite{openai2025sora2}, Wan2.6~\cite{tongyi2026wan}, Veo3, and Seedance1.5~\cite{seedance2026} can now generate coherent audio-video streams, possessing immense potential to transform the field of intelligent creation. However, T2AV, as an emerging direction, currently lacks a comprehensive and automated evaluation benchmark. Existing benchmarks, such as VBench~\cite{huang2024vbench}, primarily focus on single-modality video quality, neglecting essential audio-video synchronization. The VABench~\cite{hua2025vabench} lacks fine-grained human-related evaluations, which are critical, widely used, industry-focused, and prone to errors. It also lacks models specifically designed for the automated evaluation of AV generation.
\begin{figure}[tb]
  \centering
  \includegraphics[width=\linewidth]{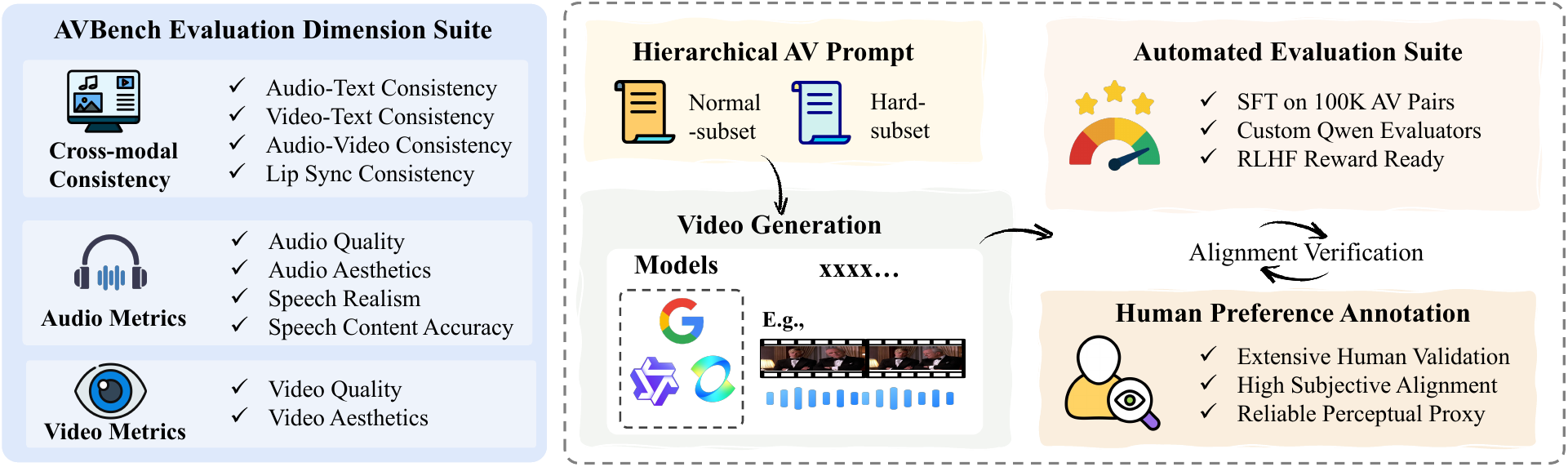}
  \caption{\textbf{Overview of our AVBench.} It integrates a multi-dimensional evaluation suite covering cross-modal consistency, audio metrics, and video metrics for human-centered real-world scenarios, together with a hierarchical AV prompt design containing normal and hard subsets. The framework supports automated large-scale assessment and human preference-based alignment verification to ensure reliable perceptual alignment.}
  \label{fig:teaser}
\end{figure}
Specifically, current evaluation methods suffer from three major deficiencies:
\textbf{1) Neglect of fine-grained human-centric evaluation:} Human-related content is the most frequently generated in T2AV applications and is a key focus for research and development companies~\cite{chen2025humo,guo2026dreamid}. Moreover, humans are the most perceptually sensitive subjects in video~\cite{yang2026human}. However, complex scenarios such as voice-character consistency in multi-person conversations, alignment of emotions, actions, and storyline, tone matching, realism of synthesized voices, lip synchronization, emotional vocal expression, and more are often overlooked in existing evaluations.
\textbf{2) The rough use of existing generic models:} Pretrained embeddings (e.g., CLAP~\cite{elizalde2023clap}, ViCLIP~\cite{wang2023viclip}) capture global semantics but lack the granularity to detect fine-grained cross-modal errors. Since these models are optimized for broad semantic matching, they often overlook subtle inconsistencies as long as the overarching audio and video themes remain conceptually consistent.
\textbf{3) Lack of precise and continuous scoring:} Mainstream evaluations rely on generic Visual Question Answering (VQA) pipelines. Crucially, existing MLLMs are not specifically trained for audio-video evaluation, and their evaluation capabilities are limited. For example, off-the-shelf MLLMs lack targeted training on fine-grained \textit{hard negatives}, making them blind to subtle cross-modal misalignments~\cite{chen2025training}. Furthermore, they typically yield discrete textual outputs rather than continuous probabilistic scores, rendering them ill-suited for direct model optimization.

To address these critical gaps, we introduce \textbf{AVBench}, a fully automated evaluation suite specifically designed for human-centric audio-video generation, as shown in Figure~\ref{fig:teaser}. First, we added detailed evaluations of human-related aspects, including cross-modal consistency, speech-related factors, and video perceptual quality in human-centered scenarios. These dimensions are carefully designed to capture the key perceptual factors in human-centered scenarios, enabling our evaluation framework to comprehensively assess a model’s capability to generate coherent, realistic, and perceptually consistent audio–video outputs in the most common human-centric settings.

In addition, we trained specialized MLLMs for the automated evaluation of AV generation. We curated 30$K$ high-fidelity human-centric real-world video clips and expanded them into a 100$K$ training set for each dimension by injecting carefully designed hard negatives, including sub-second temporal shifts, subtle emotional mismatches, and LLM-driven semantic mutations. By performing full supervised fine-tuning (SFT) on this meticulously designed dataset, the model becomes capable of detecting subtle cross-modal inconsistencies in generated content, achieving significantly higher precision than generic evaluators.
On this foundation, we fully SFT-trained 7B-parameter multimodal models to obtain three specialized consistency evaluators:  \textbf{Audio-Video (AV), Audio-Text (AT), and Video-Text (VT)}. These evaluators generate fine-grained continuous scores by normalizing the predicted probabilities of the \textit{Yes/No}  tokens. This fine-grained, differentiable output reflects the evaluator’s confidence with high precision, aligning closely with human perception. Finally, we designed a two-level evaluation pipeline with Normal and Hard subsets to test model performance in different situations. This structure helps to clearly identify weaknesses in both everyday and more challenging scenarios, giving a clearer and more complete view of the model’s generative ability.

In conclusion, our main contributions can be summarized as follows:
\begin{itemize}
    \item \textbf{Human-Centric and Comprehensive Metrics:} We have expanded upon comprehensive standard metrics by adding a wide range of fine-grained, human-centric evaluation indicators. This ensures our benchmark better meets the complex demands of real-world application scenarios.
    \item \textbf{Specialized SFT Evaluators with Continuous Scoring:}
    We developed a strategy to create negative samples and built a large, diverse training dataset across multiple dimensions. Based on this data, we trained an evaluation model that outputs continuous scores,providing a more granular and precise assessment than traditional discrete outputs. 
 \item \textbf{Automated and Human-Aligned Benchmark:} 
    AVBench uses a hierarchical structure with Normal and Hard test subsets to automatically and comprehensively evaluate model capabilities across various difficulty levels. This framework allows for a thorough assessment, specifically in core human-centric scenarios. Extensive experiments demonstrate that AVBench achieves remarkable alignment with human judgment.
    
 
\end{itemize}

\section{Related Work}
\label{sec:related}
\begin{table}[tb]
  \caption{Comparison of AVBench with existing AV generation evaluation benchmarks. Moving beyond the constraints of generic zero-shot metrics and non-automated VQA systems, AVBench establishes a specialized SFT paradigm. By leveraging 300$K$ curated pairs with hard negative mining, AVBench provides a fully automated, human-centric framework that demonstrates superior human alignment and offers a differentiable evaluation signal.}
  \label{tab:comparison}
  \centering
  \small
  \resizebox{\textwidth}{!}{%
  \begin{tabular}{@{}llccccc@{}}
    \toprule
    Benchmark & Evaluation Paradigm & Training Data & Human-Centric & Automated Eval & Human Aligned & Differentiable Signal \\
    \midrule
    VBench~\cite{huang2024vbench}       & Zero-shot Encoders & None          & \ding{55} & \ding{51}      & \ding{51} & \ding{55} \\
    T2AV-Compass~\cite{cao2025compass} & VQA-based          & None          & \ding{55} & \ding{55}      & \ding{55} & \ding{55} \\
    VABench~\cite{hua2025vabench}      & VQA-based          & None          & \ding{55} & \ding{55}      & \ding{51} & \ding{55} \\
    JointAVBench~\cite{chao2025jointavbench} & VQA-based    & None          & \ding{55} & \ding{55}      & \ding{51} & \ding{55} \\
    \midrule
    {\bf AVBench (Ours)} & {\bf Specialized SFT} & {\bf 300$K$ (w/ Hard Negs)} & {\bf \ding{51}} & {\bf \ding{51}} & {\bf \ding{51}} & {\bf \ding{51}} \\
  \bottomrule
  \end{tabular}%
  }
\end{table}

\subsection{Evolution of Native audio-video Generation}
The pursuit of high-fidelity video synthesis has evolved from silent visual generation to the unified modeling of audio-video streams. Early approaches to Text-to-audio-video (T2AV) generation predominantly adopted a cascaded paradigm, where a video is first generated by a Text-to-Video (T2V) model, followed by a separate Video-to-Audio (V2A) model conditioned on the visual frames~\cite{iashin2021taming, liu2023audioldm}. While flexible, this disjointed process often leads to temporal misalignment and semantic dissonance, as the audio generator lacks access to the latent dynamics of the video generation process.

Recent advancements have shifted toward native audio-video generation, aiming to model the joint distribution of visual and auditory modalities within a unified framework~\cite{low2025ovi,haji2025av,hayakawammdisco,wang2026klear}. This paradigm shift is driven by the scaling of Diffusion Transformers (DiT)~\cite{peebles2023scalable} and multimodal tokenizers. State-of-the-art systems such as Sora 2~\cite{openai2025sora2} and Veo 3~\cite{wiedemer2025veo3} utilize unified spatiotemporal-audio patches, enabling the simultaneous synthesis of pixel and waveform data. Similarly, Wan 2.6~\cite{tongyi2026wan} employs a hybrid attention mechanism to ensure coarse-grained semantic alignment. In the domain of human-centric generation, Seedance 1.5~\cite{seedance2026} has demonstrated establishing precise synchronization between character motion and complex audio cues (e.g., music beats or speech prosody). These native models significantly improve global coherence but introduce new challenges in fine-grained control, particularly in maintaining identity consistency and lip-sync accuracy during complex multi-talker scenarios.

\subsection{Evaluation Paradigms for Audio-Video generation}
\label{subsec:related_work_paradigms}
Table~\ref{tab:comparison} illustrates the paradigm shift in recent evaluation frameworks. Early benchmarks, such as VBench~\cite{huang2024vbench}, primarily rely on generic zero-shot embeddings (e.g., CLAP~\cite{elizalde2023clap}). While efficient in measuring global semantic similarity, these methods lack the temporal granularity necessary to detect fine-grained synchronization errors.
 However, as shown in Table~\ref{tab:comparison}, they rely on zero-shot encoders and focus on single-modality, neglecting the complex synchronization required in audio-video (AV) generation. To address cross-modal alignment, recent benchmarks such as T2AV-Compass~\cite{cao2025compass}, VABench~\cite{hua2025vabench}, and JointAVBench~\cite{chao2025jointavbench} have introduced VQA-based pipelines. While these improve human alignment, they often suffer from high inference costs and discrete outputs. Furthermore, they lack a specialized focus on human-centric scenarios, such as emotional and vocal consistency. Unlike existing post-hoc evaluators, AVBench introduces a specialized SFT framework trained on 100$K$ pairs for each evaluation dimension using hard-negative mining. It is the first to provide a differentiable evaluation signal through continuous probabilistic scoring, bridging the gap between automated assessment and active model optimization (e.g., RLHF).

\section{AVBench}
\label{sec:method}
\begin{figure}[tb]
  \centering
  \includegraphics[width=\linewidth]{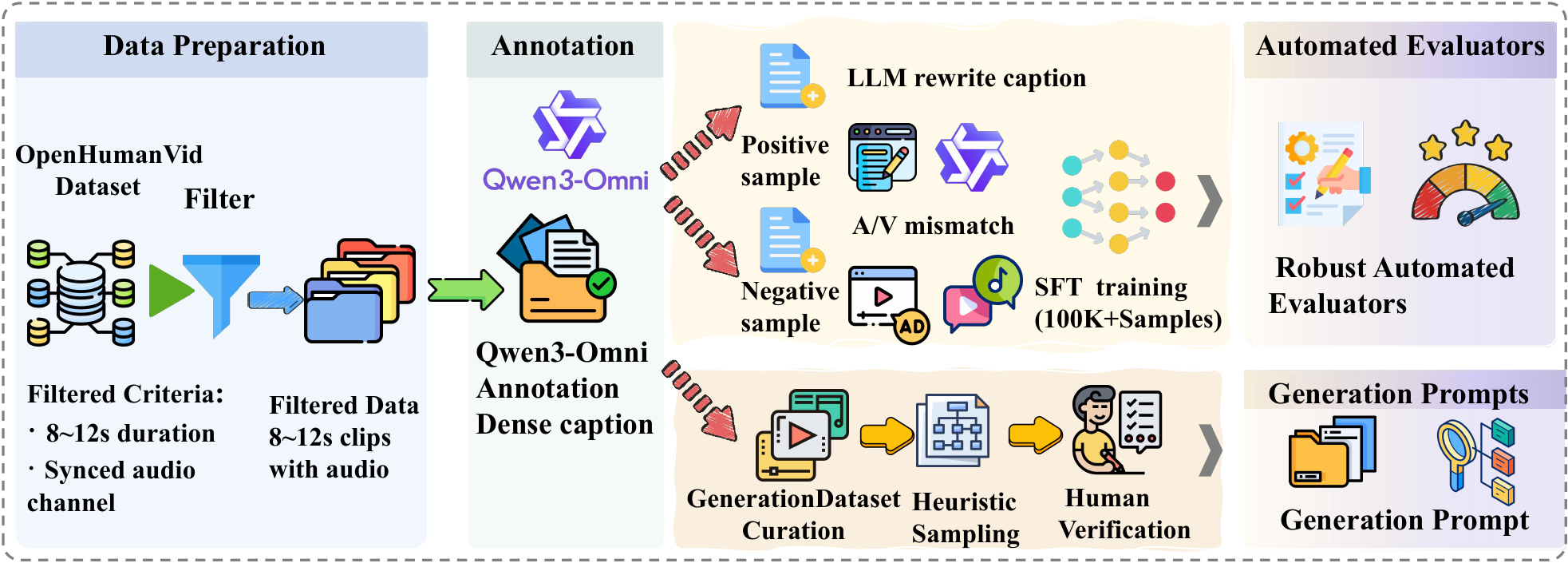}
  \caption{\textbf{Overview of the AVBench construction pipeline.} Our framework comprises two parallel workflows. The upper branch illustrates the training of the automated evaluators: 8--12\,s clips filtered from OpenHumanVid~\cite{li2025openhumanvid} are densely annotated using Qwen-Omni, and then branched into positive and hard-negative samples (via LLM rewriting and artificial A/V mismatches) for SFT. The lower branch outlines the curation of our benchmark test set: textual generation prompts are heuristically sampled from a generation dataset and strictly filtered via manual verification. Together, the robust SFT evaluators and the high-quality test prompts form our fully automated evaluation framework.}
  \label{fig:pipeline}
\end{figure}
As illustrated in Figure~\ref{fig:pipeline}, AVBench is structurally founded on two parallel pillars: (1) We developed a benchmark test set to systematically evaluate generative performance in diverse human-centric scenarios.
(2) We developed specialized automated evaluators by SFT on a novel 300$K$-sample dataset containing multi-dimensional hard negatives, which directly produce continuous and highly accurate alignment scores.

\subsection{Human-Centric Data Curation}
\label{subsec:data_curation}
AVBench employs two parallel workflows: constructing a large-scale training corpus for automated evaluators and designing a high-quality test set.

\textbf{Evaluator Training Corpus.} We extract 30$K$ real-world clips (8--12\,s) from OpenHumanVid~\cite{li2025openhumanvid} to train our evaluator models. The collected clips are predominantly human-centric, featuring people as the primary subjects.
To ensure high-fidelity semantic grounding, we utilize Qwen3-Omni~\cite{qwen3omni2026} for automated dense annotation. 
After heuristic filtering, each clip is annotated with multi-dimensional captions describing its visual, motion, and acoustic attributes. These constitute the positive samples used for evaluator SFT.

\begin{figure}[tb]
  \centering
  \includegraphics[width=0.9\linewidth]{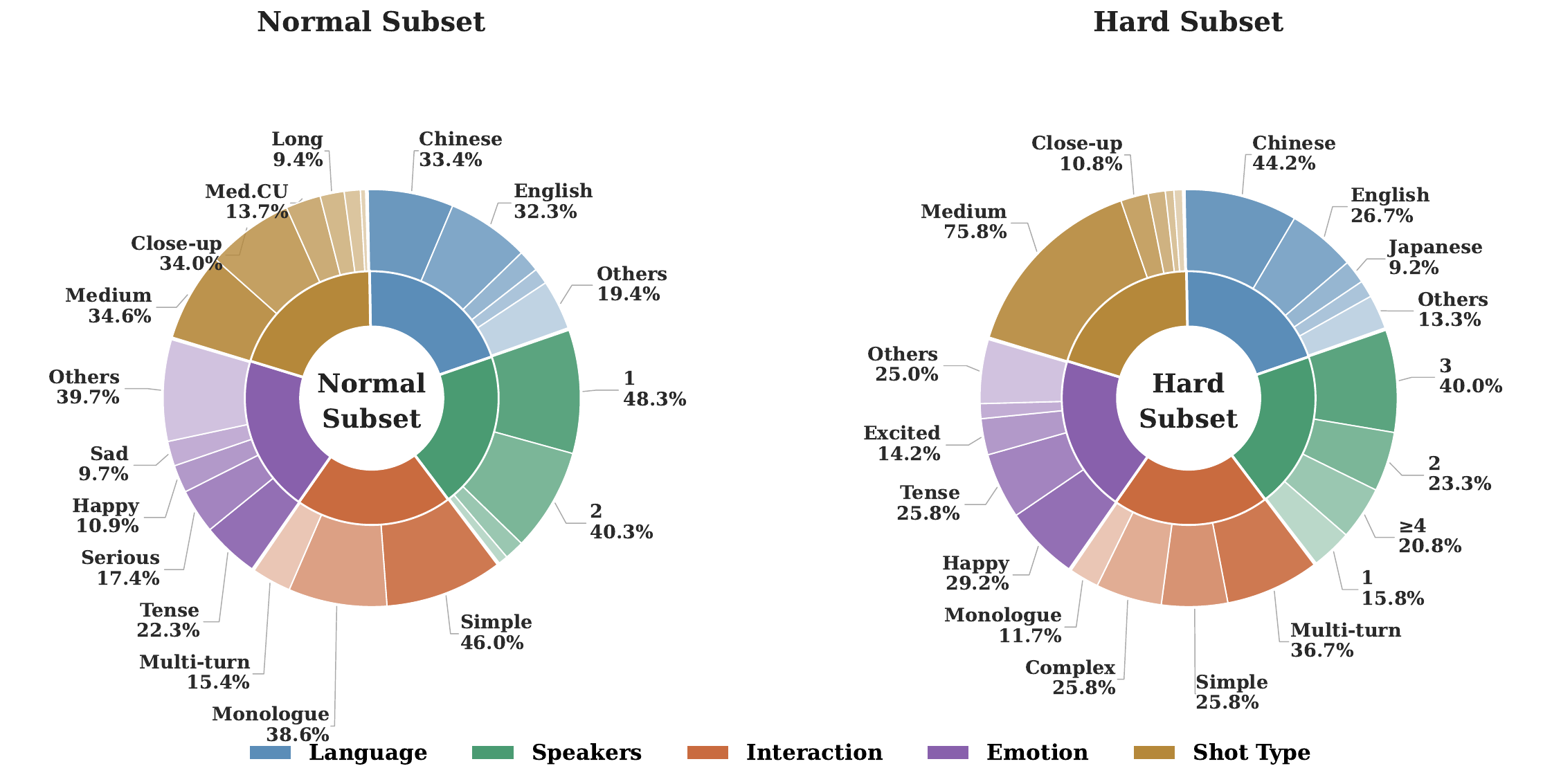}
  \caption{\textbf{Data distribution of AVBench's normal and hard subsets.} The multi-layered pie charts illustrate the comprehensive diversity of our curated evaluation dataset. The dimensions represent key human-centric attributes: \textit{Language, Number of Speakers, Interaction Complexity, Emotional Expression, and Camera Shot Type.} Notably, the hard subset incorporates a higher proportion of challenging scenarios, such as multi-turn dialogues, tense/excited emotions, and close-up shots, to rigorously test fine-grained cross-modal alignments.}
  \label{fig:data_distribution}
\end{figure}

\textbf{Test Set Curation.}
We curate 470 high-definition ($\ge 720p$) prompts from an independent pool. 
To ensure diversity, we employ a \textit{Hard Quota-Based Greedy Sampling} algorithm that enforces a 50\% upper bound on any single attribute (e.g., language, shot size). 
All prompts are manually verified for semantic clarity. 
In terms of data integrity, we adopt a strict isolation strategy to prevent leakage.
Hash-based deduplication ensures no overlap between evaluator training clips and test scenarios. Qwen3-Omni annotations are used only for SFT and are never reused as benchmark prompts. In addition, test prompts are manually rewritten to remove training-specific artifacts and ensure semantic independence.

\textbf{Hierarchical Difficulty Evaluation.} To rigorously test the performance limits and robustness of different models, we stratify the test set into two tiers based on environmental and interaction complexity (Fig.~\ref{fig:data_distribution}):
\begin{itemize}
    \item \textbf{Normal Subset:} This subset includes 1--2 speakers in stable, neutral environments with clean backgrounds, serving as a baseline test for the model's fundamental capabilities in common, everyday scenarios.
    \item \textbf{Hard Subset:} This subset is designed to test the model's robustness under challenging scenarios, including rapid/overlapping speech, noisy backgrounds, heavy occlusions, 3--4 speaker interactions, and intense emotional transitions.
\end{itemize}

\subsection{Multi-Dimensional Hard Negative Mining}
\label{subsec:negative_samples}
\begin{figure}[tb]
  \centering
  \includegraphics[width=\linewidth]{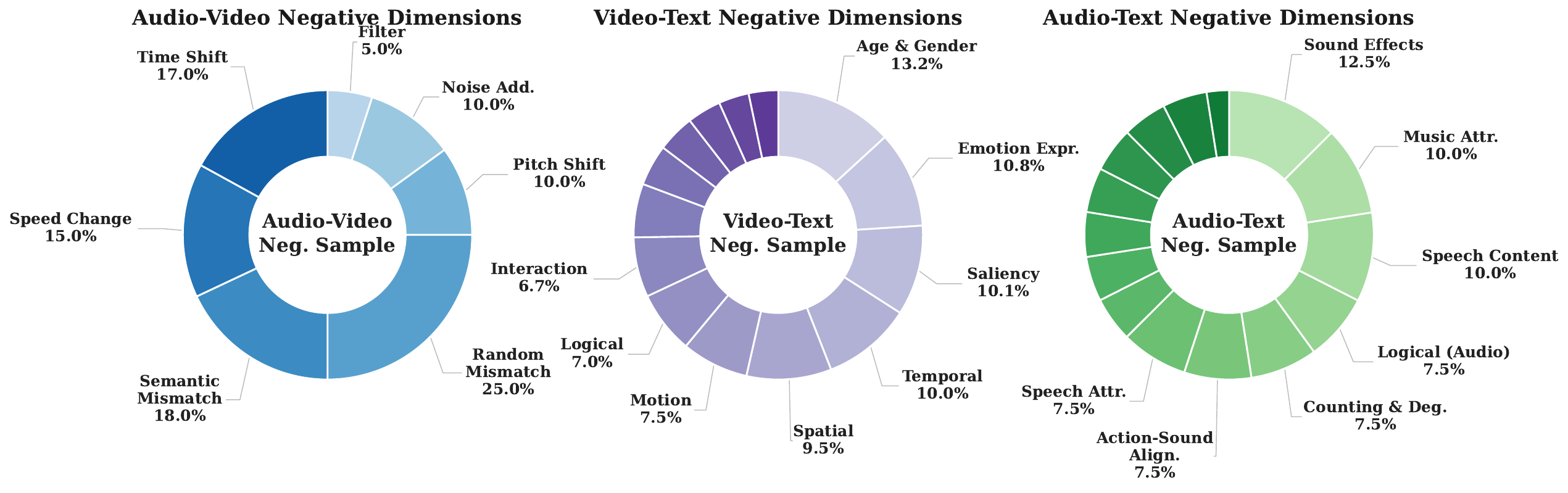}
  \caption{\textbf{Taxonomy of multi-dimensional hard negatives in AVBench.} The chart illustrates the comprehensive distribution of our constructed negative samples across three primary alignment axes. Rather than arbitrary random sampling, these fine-grained perturbation dimensions are explicitly designed to target the common failure modes of T2AV models , ensuring a robust evaluation of cross-modal consistency.}
  \label{fig:bench_dimensions}
\end{figure}

To optimize discriminative training and address the scarcity of high-quality data, we propose a systematic strategy for constructing multi-dimensional negative samples. As detailed in Figure~\ref{fig:bench_dimensions}, this approach provides extensive coverage of diverse failure modes, ranging from broad semantic shifts to fine-grained, subtle misalignments that are often imperceptible to generic models. To ensure the reliability of these complex samples, we employ an algorithmic filtering and verification pipeline that guarantees high data fidelity, providing a robust and precise foundation for sharpening the model's judgment.

\subsubsection{Automated Hard Negative Generation Pipeline.}  
To train our model on textual misalignments for Video-Text (VT) and Audio-Text (AT) modalities, we design a pipeline focusing on fine-grained, human-centric features across three stages:
1) \textbf{Dimension-Balanced Strategy.} For each sample, we select three distinct perturbation dimensions from our taxonomy. By rotating these dimensions, we ensure a balanced distribution of error types and avoid model bias.
2) \textbf{Minimal Modification via LLM.} We use the Qwen-3 Max~\cite{qwen3max2026} pipeline to make minimal changes to the original descriptions. The LLM modifies only 1–3 words while maintaining 90\%–95\% of the original structure, ensuring positive and negative pairs are similar in form but different in meaning.
3) \textbf{Algorithmic Quality Control.} We apply an automated validation filter using sequence matching to ensure the quality of the \textit{hard negatives}. Samples are accepted if their character-level similarity to the original text is within the [0.70, 0.995] range. If a sample fails, it is regenerated up to three times.
This dual-layer approach, combining LLM modifications with algorithmic validation, ensures the model focuses on meaningful cross-modal reasoning rather than surface-level word patterns.

\subsubsection{Audio-Video (AV) Misalignment:} 
To train the evaluator to effectively recognize discrepancies in physical, temporal, and environmental alignment between audio and video, we designed an automated pipeline that introduces three categories of fine-grained perturbations, each addressing common challenges in generative models. First, to rigorously test temporal synchronization, we apply Temporal Shifts, introducing micro-shifts (0.2–1.0 seconds) and medium shifts (1.0–3.0 seconds) to challenge the model’s ability to detect even subtle misalignments. Second, we simulate Acoustic Corruptions to assess physical alignment, incorporating variations in playback speed (0.8x–1.2x), pitch shifts (±2–3 semitones), and highpass/lowpass frequency filtering—typical sources of distortion in generated content. Finally, to assess Semantic and Environmental Mismatches, we utilize video metadata to introduce deliberate contradictions. These include introducing human-centric noise (e.g., overlapping speech or babbling), creating conflicts in speaker count, or mismatching ambient sounds (e.g., pairing indoor visuals with outdoor rain). By training the model on these realistic perturbations, we enable it to discern and handle complex audio-video alignment issues that are often overlooked by current generative models.

\subsubsection{Video-Text (VT) Misalignment:}
Building on our previously discussed negative sample generation pipeline, we design perturbations along key dimensions that frequently lead to errors in current audio-video generation models. These errors are critical for model performance, and by training on these targeted negative samples, our model develops a robust ability to distinguish fine-grained misalignments. Specifically, we introduce inconsistencies in \textit{Appearance}, \textit{Age \& Gender}, \textit{Emotion Expression}, and \textit{Interaction} to challenge human-centric perception. Additionally, we modify \textit{Spatial} relations, \textit{State} transitions, and \textit{Motion} trajectories, addressing common structural hallucinations in diffusion models. By training on these negative perturbations, the model is equipped with an enhanced capacity to handle the complex and subtle misalignments often overlooked by current-generation systems.

\subsubsection{Audio-Text (AT) Misalignment:}  
Our AT strategy targets semantic hallucinations and contextual discrepancies in human-centric scenarios, shifting the evaluator from simple keyword matching to more nuanced auditory reasoning. We emphasize Speech Attributes and Content (e.g., \textit{speaker\_identity}, \textit{semantic\_polarity}) to address common issues such as identity inconsistency and logical reversals, compelling the model to accurately interpret the underlying linguistic intent behind the audio.
To enhance the model's ability to detect physical inaccuracies, we manipulate Action--Sound Alignment and Sound Effects (e.g., \textit{temporal\_sync}, \textit{material\_cue}), introducing mismatches where the acoustic textures deviate from the described physical events. Additionally, perturbations in Acoustic Environment, Temporal Structure, and World Knowledge (e.g., \textit{reverberation}, \textit{physical\_plausibility}) help the model identify subtle environmental and logical inconsistencies.
These targeted training strategies equip the discriminative model with enhanced sensitivity to fine-grained cross-modal discrepancies, which are often overlooked by zero-shot encoders. By focusing on violations of both semantic coherence and physical plausibility, the model is trained to rigorously assess and distinguish between valid and invalid audio-text alignments.

\subsection{Evaluator Model Optimization via Supervised Fine-Tuning}
\label{subsec:sft}
Based on our multi-dimensional hard-negative generation strategy, we construct three balanced datasets for Audio-Text (AT), Video-Text (VT), and Audio-Video (AV) alignment, each comprising 100$K$ high-quality positive and negative pairs,totaling 300$K$ samples. We perform SFT on specialized multimodal backbones: Qwen2.5-Omni\cite{xu2025qwen25omnitechnicalreport} for the VT and AV models, and Qwen2-Audio\cite{chu2024qwen2} for the AT model.

To enhance the discriminative performance of the models, we fine-tune the LLM portion for both the AV and VT models, freezing the other components. This strategy allows us to leverage the pre-trained visual encoders and projectors, which already capture robust information representation, while focusing on adapting the LLM to internalize the alignment logic. For the AT model, we also fine-tune the connector layers in addition to the LLM. This is because the audio-text alignment involves a larger semantic gap, and training the connectors enables more flexible mapping between raw auditory features and the nuanced textual information found in the hard negatives.

During training, we employ an instruction-following template that constrains the models to generate a single token: "Yes" for alignment or "No" for misalignment. The input consists of the multimodal content paired with a task-specific query (e.g., \textit{"Does the audio accurately match the video content? Answer only Yes or No."}). By supervising the models to concentrate probability mass on these two specific tokens, we derive a normalized scoring signal \(S\), which is defined as the ratio of the probability of the model outputting "Yes" to the sum of the probabilities of outputting "Yes" and "No". This differentiable signal provides a continuous evaluation score, allowing for a more accurate assessment of the alignment between different modalities and serving as a reliable metric for model performance.

\subsection{The Comprehensive Evaluation Suite}
\label{subsec:Evaluation Suite}
To provide a more comprehensive and precise evaluation, \textbf{AVBench} enhances our SFT evaluators by incorporating lip-sync analysis and unimodal quality metrics. This integration bridges the gap between high-level semantic alignment and low-level signal fidelity, ensuring a thorough assessment of both physical realism and structural integrity. The framework is divided into two modules: Cross-Modal Alignment \& Synchronization and Unimodal Generation Quality, which encompass 10 distinct dimensions to fully capture and reflect the diverse capabilities of audio-video generation models.

\subsubsection{Cross-Modal Alignment \& Synchronization}
This module assesses both the semantic alignment and temporal synchronization between different modalities, which are key challenges for current native T2AV models. It focuses on the following four dimensions:
\begin{itemize}
    \item \textbf{AT, VT, and AV Consistency:} We leverage our specialized SFT-optimized evaluators to assess alignment across the Audio-Text, Video-Text, and Audio-Video axes. By capturing fine-grained semantic and temporal dependencies, these models provide a highly accurate and comprehensive measure of cross-modal consistency, effectively identifying subtle misalignments that traditional metrics often overlook.
    \item \textbf{Lip Sync Consistency:} A critical component of human-centric generation is the precise physical alignment between mouth movements and emitted speech. To evaluate this, we utilize the SyncNet architecture~\cite{chung2016out} implemented via the LatentSync framework~\cite{latentsync2024}. By integrating alignment confidence with temporal offset analysis, we derive a holistic fidelity score that effectively reflects the accuracy of lip-sync alignment in the generated video, ensuring a more comprehensive and systematic evaluation of cross-modal alignment.
\end{itemize}

\subsubsection{Unimodal Generation Quality}
This module rigorously assesses the signal-level fidelity, physical realism, and overall aesthetics of the generated audio and video streams independently across the remaining six dimensions:
\begin{itemize}
    \item \textbf{Speech Content Accuracy:} current audio-video generation models frequently suffer from nonsensical vocalizations or fail to faithfully reflect specified textual prompts, making this dimension crucial for verifying the accuracy of spoken content. We utilize \textit{Whisper-large-v3}~\cite{radford2023robust} for transcription and compute a comprehensive final score by weighting three specific components: keyword completeness ($S_{comp}$), lexical accuracy ($S_{acc}$), and a hallucination penalty($S_{hall}$) for extraneous content. The final weighted aggregation ensures that the benchmark captures both missing information and spurious auditory hallucinations.
    \item \textbf{Speech Realism:}
    Generated speech should exhibit natural prosody, coherent intonation, and realistic timbral characteristics. To assess voice-cloning fidelity and overall speech naturalness, we employ \textit{DF\_Arena}~\cite{df_arena}, a high-capacity discriminator specifically designed to differentiate authentic human speech from synthetic generative artifacts.
    \item \textbf{Audio Quality:}
    To comprehensively evaluate overall audio quality across both speech and natural acoustic signals, we employ the NISQAv2 metric~\cite{mittag2021nisqa}. In our assessment framework, the mean opinion score (MOS) predicted by NISQAv2 is adopted as the principal indicator of perceptual audio quality, as it provides a holistic measure of overall speech quality.
    \item \textbf{Audio Aesthetics:} To measure the aesthetic value of audio, we employ the \textit{Audiobox-Aesthetics}~\cite{vyas2023audiobox} module, which holistically assesses audio production quality across four key dimensions: Content Enjoyment (CE), Content Usefulness (CU), Production Complexity (PC), and Production Quality (PQ). Following the evaluation framework in~\cite{shan2025hunyuanvideo}, which suggests that production complexity is inversely correlated with perceived quality, we define our aggregated audio aesthetic score as $(CE + CU + PQ - PC) / 4$. This score provides a balanced metric to quantify the overall aesthetic value of the audio production.
    \item \textbf{Video Quality:} To provide a comprehensive assessment of the generated videos, we utilize \textit{DOVER++}~\cite{wu2023dover++}. As a state-of-the-art multi-perspective metric, it generates a holistic technical score that accurately reflects the overall quality and signal-level fidelity of the video content.
    \item \textbf{Video Aesthetics:} We utilize the \textit{LAION-Aesthetics} predictor~\cite{schuhmann2022laion} to assess the overall artistic composition and visual appeal of the generated content. This metric provides a robust measure of high-level aesthetic quality, offering a comprehensive appraisal of the video's overall aesthetic value.
\end{itemize}

By unifying these ten fine-grained dimensions, \textbf{AVBench} establishes a rigorous, multi-faceted evaluation framework.

\section{Experiments and Analysis}
\label{sec:experiments}

\subsection{Main Results: Evaluating State-of-the-Art T2AV Models}
\label{subsec:main_results}
\begin{figure}[tb]
  \centering
  \begin{subfigure}{0.4\linewidth}
    \centering
    \includegraphics[width=\linewidth]{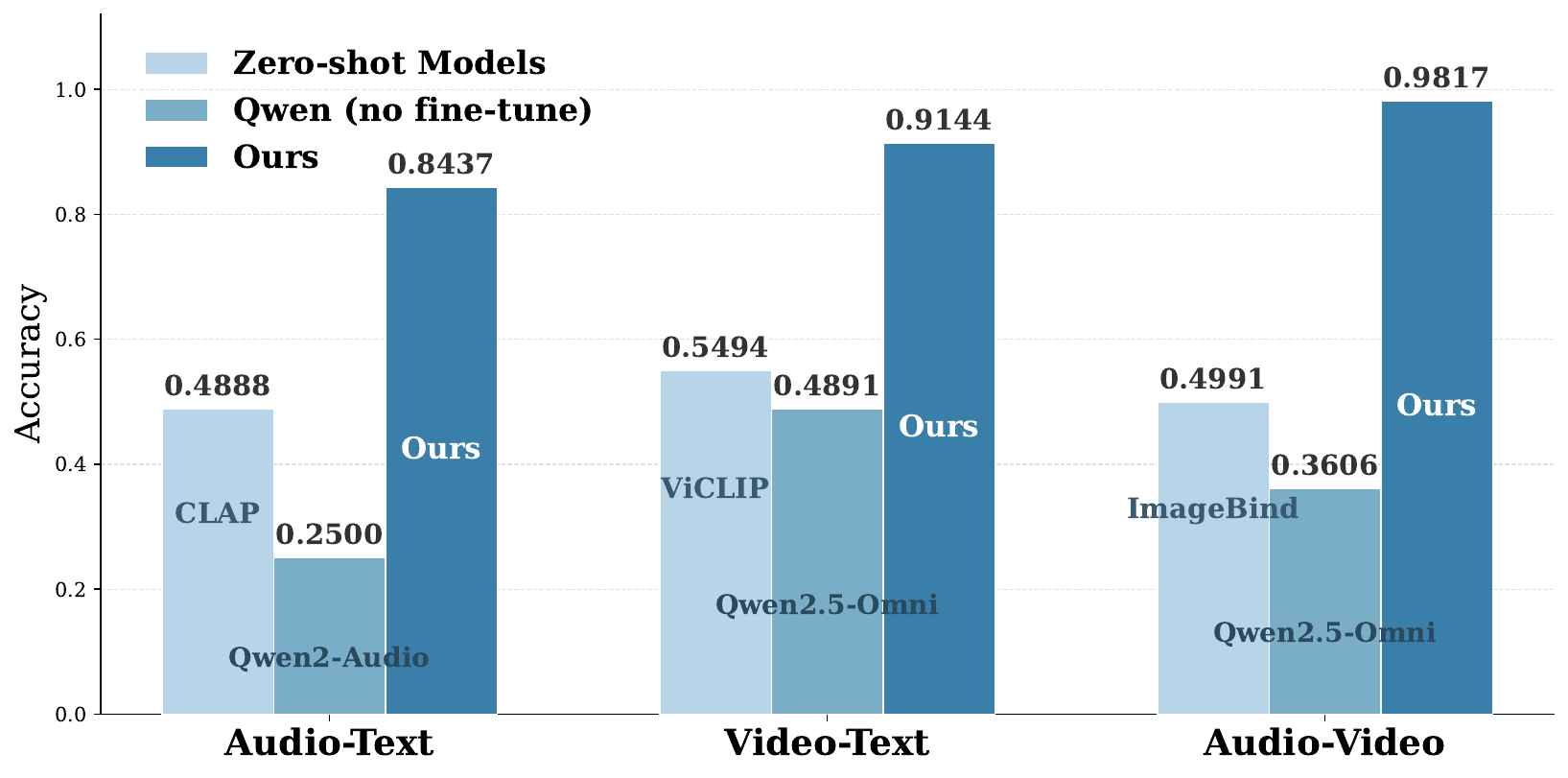} 
    \caption{Evaluator Reliability}
    \label{fig:evaluator_accuracy}
  \end{subfigure}
  \hfill
  \begin{subfigure}{0.58\linewidth}
    \centering
    \includegraphics[width=\linewidth]{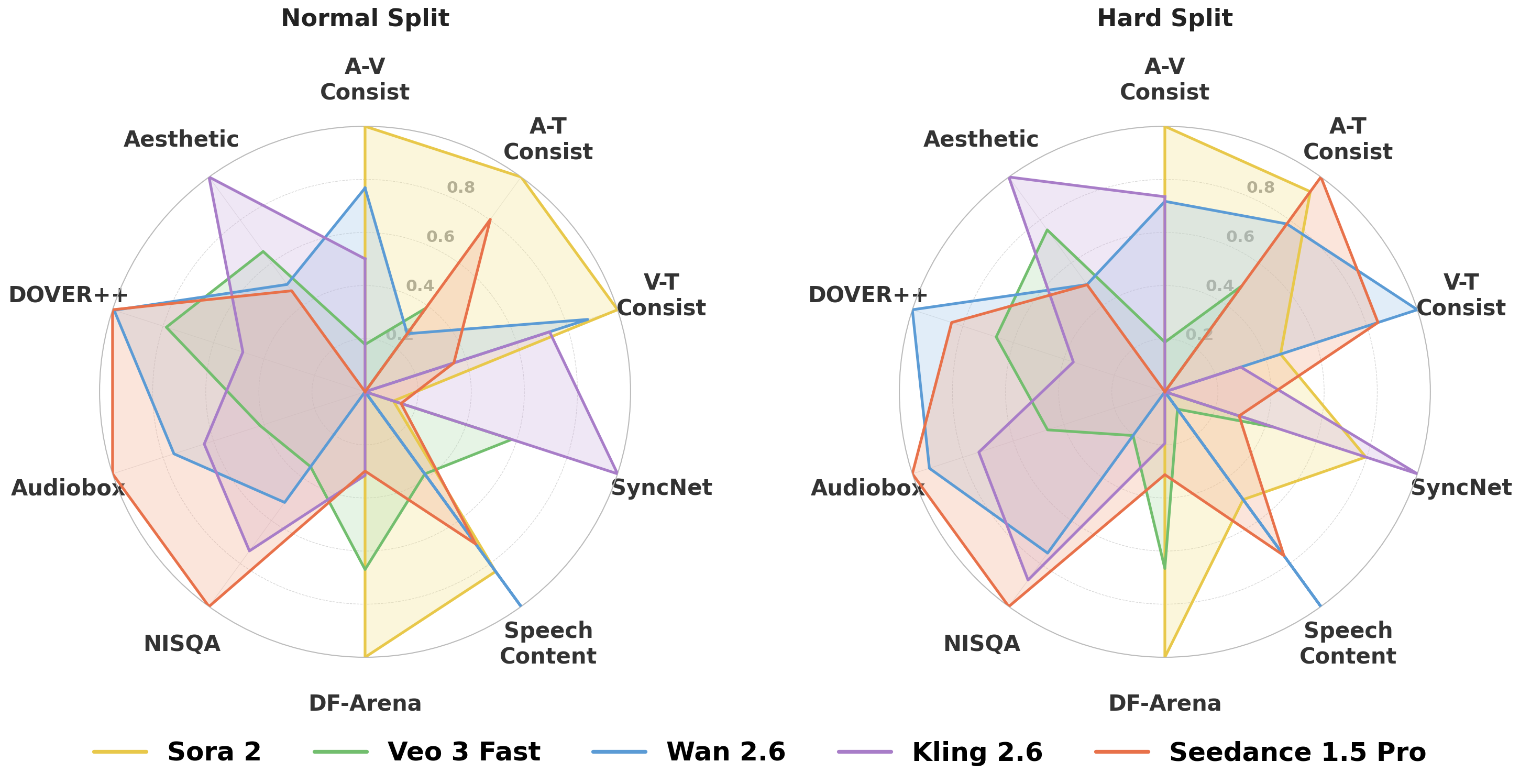} 
    \caption{Holistic Model Performance}
    \label{fig:radar_chart}
  \end{subfigure}
  \caption{\textbf{Comprehensive evaluation framework and model benchmarking.} 
  (a) \textbf{Detection accuracy on hard-negative test sets}: A comparison between our specialized SFT evaluators and generic MLLM baselines. 
    (b) \textbf{Multi-dimensional evaluation radar chart}: A holistic profile of five T2AV models across Normal (left) and Hard (right) splits. Metrics are independently Min-Max normalized $[0, 1]$ within each split ($1.0$ indicates the best).}
  \label{fig:main_results_visualization}
\end{figure}

Based on the data in Table \ref{tab:main_results} and Figure \ref{fig:radar_chart}, the evaluation reveals clear performance differences among the five T2AV models, particularly under the Hard Split conditions.

\textbf{Cross-Modal Alignment \& Synchronization.}

The data identifies Video-Text (VT) consistency as the main bottleneck for all models. In the Normal Split, Sora 2 achieves an AT score of 0.8675 but a lower VT score of 0.7599. While Sora 2 improves its AV score to 0.9320 in the Hard Split, its VT score drops to 0.7190. This suggests that as prompts become more complex, models prioritize audio-video synchronization over strict adherence to detailed text instructions. Across all models, VT scores consistently lag behind AV scores, indicating that following text instructions in the visual modality remains a major challenge.

\textbf{Lip-Sync, Speech Content, \& Realism.}
Results show a clear decoupling between synchronization, content accuracy, and vocal realism. Kling 2.6 achieves the best lip-sync performance ($8.1027$ in Normal, $3.9844$ in Hard) but performs weaker in speech content accuracy and realism. Wan 2.6 attains the highest SC scores ($91.5568$ in Normal, $84.4512$ in Hard) despite having the lowest vocal realism. Sora 2 presents a more balanced profile, producing the most natural voices while maintaining competitive content accuracy across both splits.

\textbf{Technical Quality and Aesthetics.}
Seedance 1.5 Pro leads in technical quality, achieving the highest scores in audio quality (NISQA), audio aesthetics (Audiobox), and video technical quality (DOVER++), with consistently strong results in the Hard Split. In contrast, Kling 2.6 produces the most visually appealing outputs, obtaining the highest aesthetic scores across both splits.

\textbf{Summary of Model Trends.}

The results highlight that T2AV models struggle with precise text instruction following, particularly in VT consistency, which remains a notable bottleneck across all models. This issue is exacerbated in the Hard Split, showing that current systems cannot yet fully handle complex, human-centric scenarios. While models often excel in technical quality and aesthetics , these surface-level strengths do not compensate for the lack of realism and semantic depth required for high-fidelity generation.

\begin{table}[tb]
  \caption{Quantitative evaluation of state-of-the-art T2AV models on the AVBench test set across normal and hard splits. We report \textbf{Cross-Modal Alignment \& Synchronization} via specialized SFT evaluators (AV, AT, and VT Consistency) and Lip Sync Consistency (SyncNet). \textbf{Unimodal Generation Quality} is assessed through speech content accuracy (SC), speech realism (DF-Arena), audio quality (NISQA MOS), audio aesthetics (Audiobox), video technical quality (DOVER++), and video aesthetics. $\uparrow$ indicates higher is better.}
  \label{tab:main_results}
  \centering
  \resizebox{\textwidth}{!}{%
  \begin{tabular}{@{}l ccc c cccccc@{}}
    \toprule
    & \multicolumn{4}{c}{Cross-Modal Alignment \& Sync} & \multicolumn{6}{c}{Unimodal Generation Quality} \\
    \cmidrule(lr){2-5} \cmidrule(lr){6-11}
    Model & AV $\uparrow$ & AT $\uparrow$ & VT $\uparrow$ & SyncNet $\uparrow$ & SC $\uparrow$ & DF-Arena $\uparrow$ & NISQA $\uparrow$ & Audiobox $\uparrow$ & DOVER++ $\uparrow$ & Aesthetic $\uparrow$ \\
    \midrule
    \rowcolor[HTML]{F2F2F2} \multicolumn{11}{l}{\textit{Normal Split}} \\
    Sora 2            & \textbf{0.8713} & \textbf{0.8675} & \textbf{0.7599} & 4.9057          & 87.8391          & 0.4328          & 2.3784          & 3.1759          & 60.0125          & 4.0704          \\
    Veo 3 Fast        & 0.6924          & 0.8300          & 0.7235          & 6.5943          & 77.4950          & 0.3043          & 2.8191          & 3.5877          & 69.2275          & 4.9967          \\
    Wan 2.6           & 0.8207          & 0.8227          & 0.7556          & 4.5016          & \textbf{91.5568} & 0.0441          & 3.0289          & 3.9271          & 71.6473          & 4.7790          \\
    Kling 2.6         & 0.7626          & 0.8061          & 0.7501          & \textbf{8.1027} & 68.7844          & 0.1665          & 3.3141          & 3.8082          & 65.6786          & \textbf{5.4885} \\
    Seedance 1.5 Pro  & 0.6536          & 0.8554          & 0.7363          & 5.0146          & 84.9268          & 0.1602          & \textbf{3.6411} & \textbf{4.1686} & \textbf{71.7205} & 4.7373          \\
    \midrule
    \rowcolor[HTML]{F2F2F2} \multicolumn{11}{l}{\textit{Hard Split}} \\
    Sora 2            & \textbf{0.9320} & 0.8575          & 0.7190          & 3.7932          & 76.7905          & \textbf{0.5498} & 2.0564          & 3.1339          & 58.1538          & 4.0434          \\
    Veo 3 Fast        & 0.7766          & 0.8117          & 0.6943          & 3.4535          & 70.3144          & 0.3827          & 2.3321          & 3.6113          & 67.0833          & 5.1438          \\
    Wan 2.6           & 0.8780          & 0.8418          & \textbf{0.7482} & 3.0488          & \textbf{84.4512} & 0.0498          & 3.0726          & 4.0924          & \textbf{71.5229} & 4.7721          \\
    Kling 2.6         & 0.8813          & 0.7602          & 0.7105          & \textbf{3.9844} & 69.0691          & 0.1469          & 3.2425          & 3.8912          & 62.9994          & \textbf{5.5033} \\
    Seedance 1.5 Pro  & 0.7409          & \textbf{0.8646} & 0.7398          & 3.3239          & 80.8029          & 0.2059          & \textbf{3.4093} & \textbf{4.1618} & 69.4430          & 4.7707          \\
    \bottomrule
  \end{tabular}%
  }
\end{table}

\begin{figure}[tb]
  \centering
  \includegraphics[width=\linewidth]{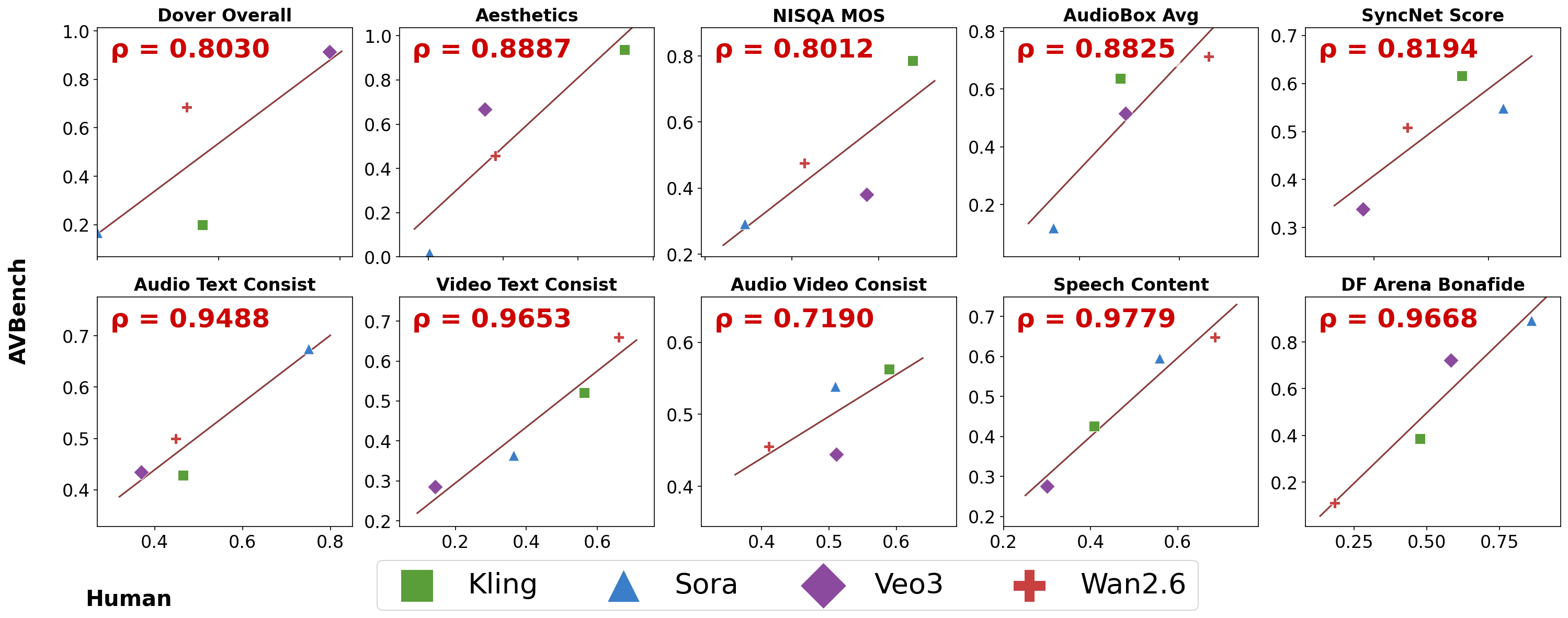}
  \caption{\textbf{Pearson Correlation between AVBench automated scores and human preference.}
  Each subplot corresponds to one of the 10 evaluation dimensions.
  The $x$-axis denotes the human win ratio per model,
  and the $y$-axis shows metric win ratio. Each point represents one model family
  (Kling~2.6, Sora~2, Veo~3 Fast, Wan~2.6),
  and the red line shows the least-squares linear fit.}
  \label{fig:pearson_correlation}
\end{figure}

\subsection{Effectiveness of the SFT Evaluator}
\label{subsec:sft_effectiveness}
To demonstrate the necessity of our SFT paradigm, we compare our specialized evaluators against zero-shot foundational models (CLAP, ViCLIP, ImageBind~\cite{girdhar2023imagebind}) and non-finetuned MLLMs on hard-negative test sets.

As illustrated in Figure~\ref{fig:evaluator_accuracy}, zero-shot encoders struggle significantly with fine-grained discrimination, often hovering around random chance. For instance, CLAP yields only 0.4888 accuracy on Audio-Text, and ImageBind achieves 0.4991 on Audio-Video alignment. Similarly, directly prompting base MLLMs without SFT results in poor performance due to a severe positive bias. These models tend to default to \textit{Yes} and habitually output positive judgments regardless of the actual alignment. Qwen2-Audio, for example, achieves only 0.2500 on Audio-Text consistency.

Conversely, by training on 100K curated pairs with hard negatives, the discriminative capability of our specialized SFT evaluators is significantly enhanced. We achieve high accuracies across all dimensions: 0.8437 for Audio-Text, 0.9144 for Video-Text, and a peak of 0.9817 for Audio-Video consistency. These results demonstrate that targeted fine-tuning is essential for detecting the specific flaws in native T2AV generation.

\subsection{Human Alignment Validation}
\label{subsec:human_alignment}
To verify AVBench's reliability, we conduct a Pearson correlation analysis between human win ratios and automated scores (Figure~\ref{fig:pearson_correlation}). Four domain experts performed pairwise comparisons (2AFC) on videos generated from the same prompt, selecting the superior model per dimension with a tie option. The model-level win ratio is computed as
\begin{equation}
\mathrm{WinRatio} = \frac{W + 0.5\,T}{W + T + L},
\end{equation}
where $W$, $T$, and $L$ denote the numbers of wins, ties, and losses, respectively. 
Both the human preference axis and the automated metric axis adopt this pairwise win-ratio formulation.
Pearson correlation is then computed at the model level between the two resulting win-ratio sequences using the standard  Pearson correlation coefficient.

As shown in Figure~\ref{fig:pearson_correlation}, AVBench exhibits strong human alignment. Our SFT-trained evaluators achieve high correlation for AT Consistency ($\rho = 0.9488$) and VT Consistency ($\rho = 0.9653$). Speech realism and quality metrics also align closely, notably Speech Content ($\rho = 0.9779$), DF Arena Bonafide ($\rho = 0.9668$), and NISQA MOS ($\rho = 0.8012$). Furthermore, the SyncNet Score exhibits a strong correlation ($\rho = 0.8194$). These findings highlight AVBench as a robust and reliable evaluation framework for characterizing the comprehensive capabilities of audio-video generation models, providing a high-fidelity and fine-grained assessment signal for audio-video quality and consistency.

\section{Limitation}
\label{sec:limitation}
Despite providing a comprehensive and automated evaluation framework, AVBench currently focuses primarily on short-form video clips of 5–12\,s, which aligns with the default output lengths of state-of-the-art models such as Sora 2 and Veo 3. In future work, the same framework can be naturally extended to construct training data and evaluation benchmarks for longer videos, enabling systematic assessment of longer-duration video generation using the same methodology.

\section{Conclusion}
\label{sec:conclusion}
In this paper, we introduced \textbf{AVBench}, a fully automated evaluation framework specifically designed for human-centric audio-video generation. By utilizing a specialized SFT pipeline trained on 300$K$ samples with multi-dimensional hard negatives, the framework effectively captures subtle cross-modal misalignments that traditional metrics often overlook. AVBench provides a comprehensive assessment of model capabilities across ten distinct dimensions, integrating semantic consistency, signal-level quality, and aesthetics. Furthermore, by deriving continuous, differentiable scores from model confidence, our framework demonstrates significant potential as a scalable reward signal for reinforcement learning-based alignment. Validated by its strong correlation with human judgment, AVBench serves as a robust proxy for the systematic evaluation and optimization of audio-video generation systems.



%
%
\bibliographystyle{splncs04}
\bibliography{main}
\newpage
\section{Extended Details on Negative Sample Construction}
\label{sec:intro}
\subsection{Negative Sample Construction Pipeline}
To systematically evaluate the fine-grained alignment capabilities of Text-to-Audio-Video (T2AV) models, we employ a rigorous negative sample construction pipeline. Specifically, we utilize Large Language Models (LLMs) to modify the original textual descriptions, intentionally introducing fine-grained mismatches between text and video, as well as between text and audio.

\begin{figure}[h]
  \centering
  \includegraphics[width=\linewidth]{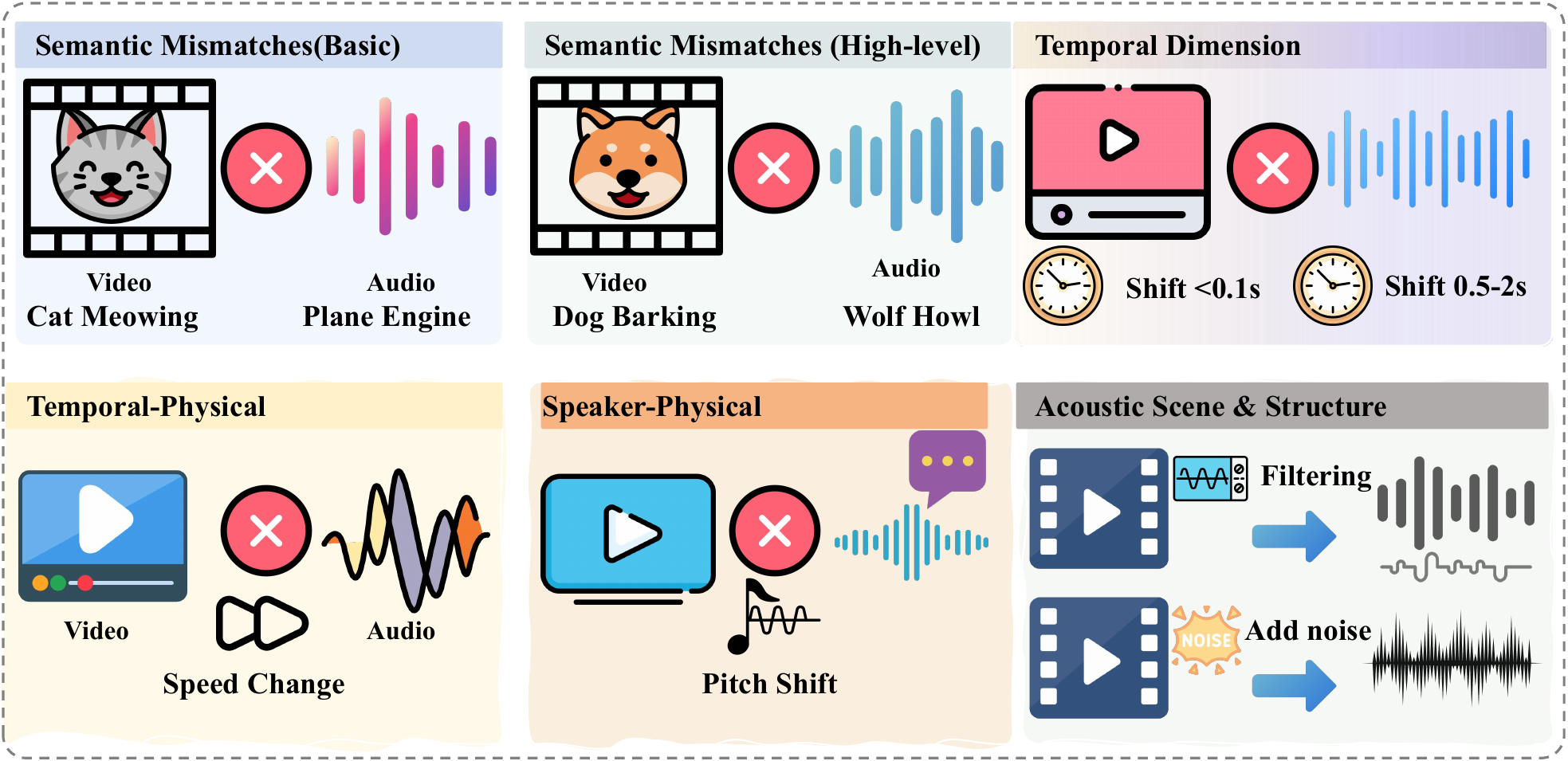}
  \caption{Overview of audio-video negative sample construction across diverse dimension levels. The strategies include \textbf{Basic} and \textbf{High-level Semantic Negatives} via \textit{random\_mismatch} and \textit{semantic\_mismatch}, alongside \textbf{High-precision Temporal} and \textbf{Temporal Negatives} utilizing micro and medium time shifts. Furthermore, physical and acoustic properties are perturbed to generate \textbf{Temporal-Physical} and \textbf{Speaker/Physical Negatives} (\textit{speed\_change}, \textit{pitch\_shift}), as well as \textbf{Acoustic Scene} and \textbf{Acoustic Structural Negatives} (\textit{noise\_addition}, \textit{filter}).}
  \label{fig:audio_video}
\end{figure}

\textbf{Audio-Video Mismatch Construction:} As illustrated in Fig. \ref{fig:audio_video} and detailed in Table \ref{tab:av_negative_samples}, negative samples are generated by manipulating video and audio tracks across eight specific dimension levels. Specifically, to construct semantic conflicts, we utilize \textit{random\_mismatch} (e.g., Cat vs. Plane) to create a \textbf{Basic Semantic Negative}, and \textit{semantic\_mismatch} (e.g., Dog vs. Wolf) for a \textbf{High-level Semantic Negative}. For temporal and physical synchronizations, we employ \textit{time\_shift\_micro} ($<0.1s$) to generate a \textbf{High-precision Temporal Negative}, and \textit{time\_shift\_medium} ($0.5$--$2s$) for a standard \textbf{Temporal Negative}. Additionally, \textit{speed\_change} is applied to introduce a \textbf{Temporal-Physical Negative}. Auditory identity attributes are perturbed via \textit{pitch\_shift} to construct a \textbf{Speaker/Physical Negative}. Finally, acoustic environment consistency is disrupted using \textit{noise\_addition} to generate an \textbf{Acoustic Scene Negative}, and spectral \textit{filter} (e.g., low-pass) to create an \textbf{Acoustic Structural Negative}.

\textbf{Audio-Text and Video-Text Mismatch Construction:} For cross-modal text consistency, we leverage Large Language Models to systematically perturb the original ground-truth captions based on our predefined dimensions. To effectively capture the distinct characteristics of different modalities, we design specific prompt templates that instruct the LLM to alter target attributes while keeping the rest of the description logically coherent.

\begin{figure}[p]
  \centering
  \includegraphics[width=\linewidth]{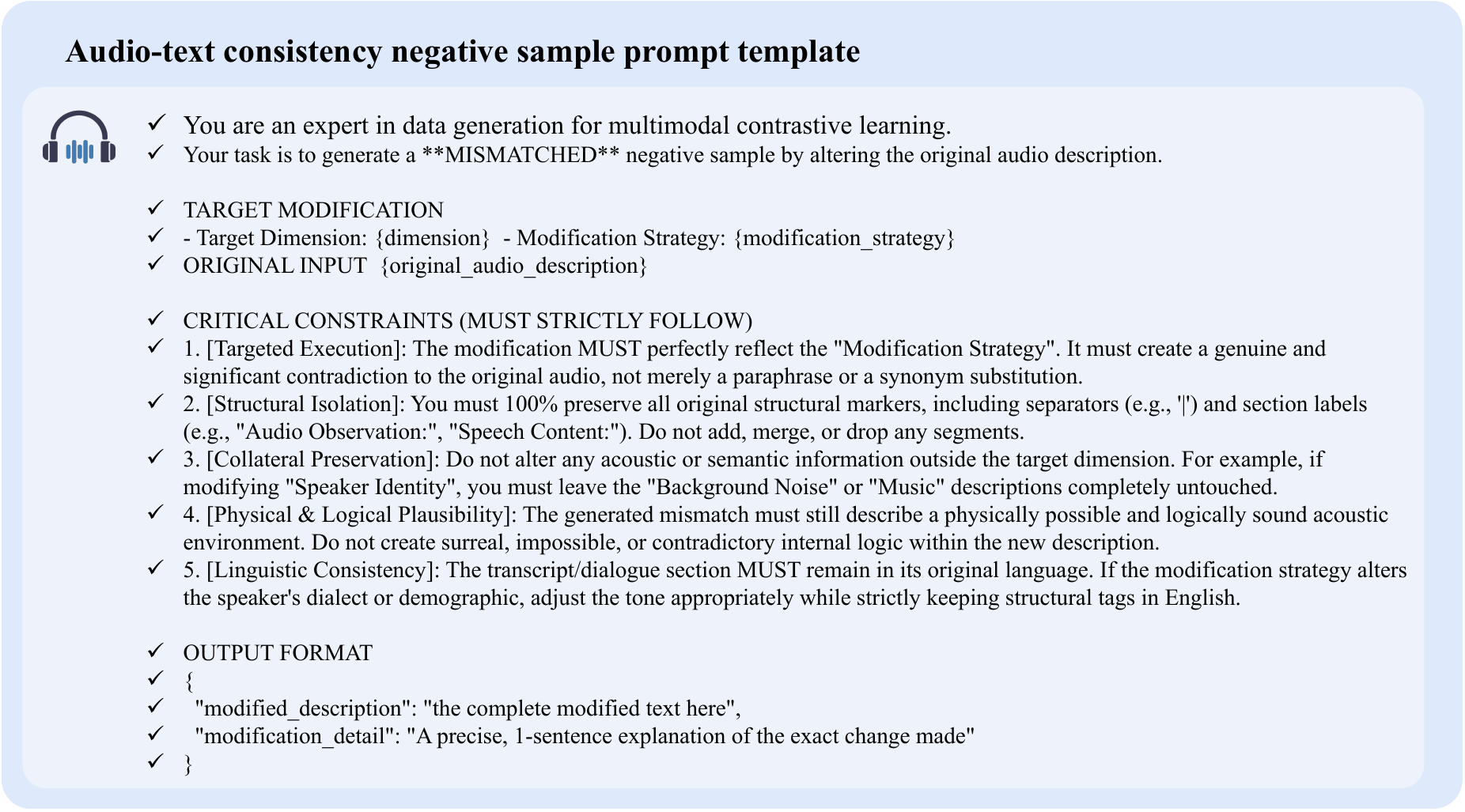}
  \caption{Prompt template for generating audio-text consistency negative samples. The template instructs the model to construct mismatched descriptions for multimodal contrastive learning by enforcing critical constraints such as targeted execution, structural isolation, and collateral preservation.}
  \label{fig:audio_text_prompt}
\end{figure}

For audio-text mismatch construction, the prompt instructs the model to generate a "MISMATCHED" negative sample by altering the original audio description. As shown in Figure \ref{fig:audio_text_prompt}, this process is governed by critical constraints such as "Structural Isolation," which requires the LLM to 100\% preserve all original structural markers and section labels. Furthermore, "Collateral Preservation" ensures that no acoustic or semantic information outside the target dimension is altered, and the resulting description must maintain "Physical \& Logical Plausibility" to avoid surreal or contradictory internal logic.

\begin{figure}[p]
  \centering
  \includegraphics[width=\linewidth]{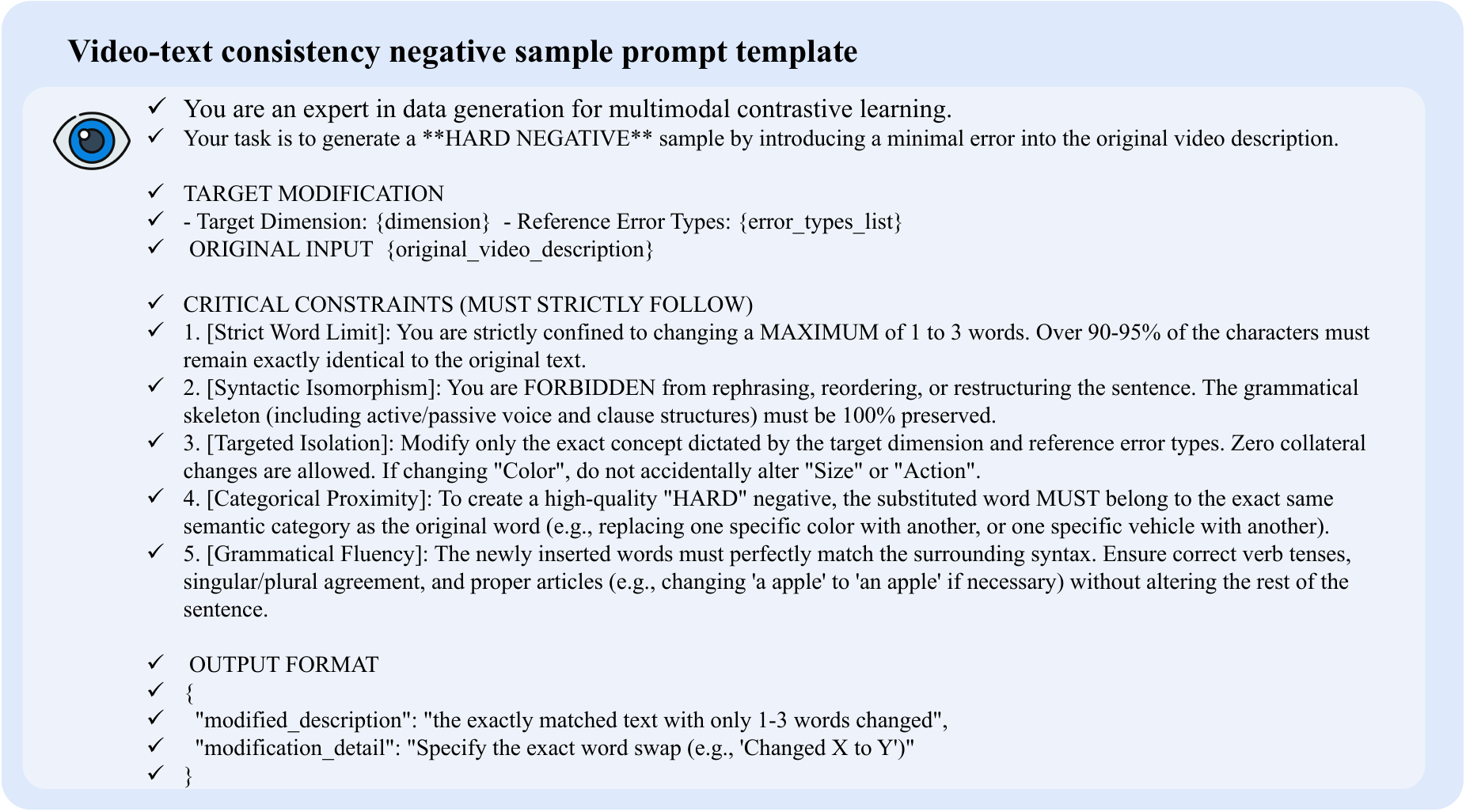}
  \caption{Prompt template for generating video-text consistency negative samples. The template instructs the model to construct hard negative samples by introducing a minimal error into the original description, strictly guided by constraints such as syntactic isomorphism and categorical proximity.}
  \label{fig:video_text_prompt}
\end{figure}

Conversely, for video–text mismatch construction, the prompt is designed to generate hard negative samples by introducing minimal yet critical errors into the original video descriptions. As illustrated in Figure \ref{fig:video_text_prompt}, this is achieved through a Strict Word Limit, which constrains modifications to at most 1–3 words. To prevent the LLM from introducing trivial variations, the prompt further enforces Syntactic Isomorphism, prohibiting any rephrasing or structural changes so that the original grammatical structure is strictly preserved. In addition, Categorical Proximity requires that substituted words belong to the same semantic category as the original terms, ensuring that the resulting mismatches remain subtle and semantically plausible.

We execute these modality-specific prompts across all secondary dimensions defined in our taxonomy, followed by a manual verification phase to ensure the generated negatives do not inadvertently become false positives.

\subsection{Detailed Taxonomy of Negative Samples}
Our benchmark categorizes multimodal inconsistencies into highly granular dimensions to probe the exact weaknesses of current foundation models. 

Table~\ref{tab:av_negative_samples} details the taxonomy for audio-video consistency, focusing heavily on temporal, and semantic synchronization. Table~\ref{tab:audiotext_negative_samples} provides a comprehensive hierarchy for audio-text inconsistencies, ranging from speech attributes to high-level cultural cues. Finally, Table~\ref{tab:videotext_negative_samples} outlines the video-text dimensions, covering aspects from visual appearance to complex logical reasoning.

Regarding the data distribution, we ensure a balanced representation across the primary categories to prevent evaluation bias. However, slightly more weight (approx. 25\%) is allocated to temporal and semantic alignments, as these are the most fundamental capabilities required for robust multimodal understanding. The remaining samples are uniformly distributed among fine-grained dimensions such as counting, state, and spatial relations.

\begin{table}[p]
  \caption{A detailed taxonomy of audio-video consistency negative samples, mapping conceptual dimension levels to their specific manipulation strategies and code implementations.}
  \label{tab:av_negative_samples}
  \centering
  \begin{tabular}{@{} l l l @{}}
    \toprule
    Dimension Level & Main Strategy & Code Strategy \\
    \midrule
    {\bf Basic Semantic Negative} & Random Mismatch & \textit{random\_mismatch} \\
    {\bf High-level Semantic Negative} & Semantic Mismatch & \textit{semantic\_mismatch} \\
    \noalign{\vspace{1ex}}
    {\bf High-precision Temporal Negative} & Micro Time Shift & \textit{time\_shift\_micro} \\
    {\bf Temporal Negative} & Medium Time Shift & \textit{time\_shift\_medium} \\
    {\bf Temporal-Physical Negative} & Speed Change & \textit{speed\_change} \\
    \noalign{\vspace{1ex}}
    {\bf Speaker/Physical Negative} & Pitch Shift & \textit{pitch\_shift} \\
    \noalign{\vspace{1ex}}
    {\bf Acoustic Scene Negative} & Noise Addition & \textit{noise\_addition} \\
    {\bf Acoustic Structural Negative} & Spectral Filtering & \textit{filter} \\
    \bottomrule
  \end{tabular}
\end{table}

\begin{table}[p]
  \caption{A comprehensive taxonomy of video-text consistency negative samples, outlining the evaluated dimensions, core focus areas, and typical error types generated for mismatch construction.}
  \label{tab:videotext_negative_samples}
  \centering
  \small 
  \begin{tabular}{@{} p{3.2cm} p{3.5cm} p{5.0cm} @{}}
    \toprule
    Dimension & Core Focus & Typical Error Types \\
    \midrule
    {\bf Appearance} & Visual attributes & Color, shape, material, clothing, details \\
    {\bf Emotion \& Expression} & Emotional expressions & Emotion, facial expression, body language \\
    {\bf Age \& Gender} & Age and gender & Age change, gender swap \\
    {\bf Social Relation} & Social relationships & Relationship type, intimacy, interaction context \\
    {\bf Counting} & Quantity & Number of people/objects, object existence \\
    {\bf Motion} & Movement and actions & Action, phase, trajectory, speed, frequency \\
    {\bf Interaction} & Interactions & Human-object, object-object relationships \\
    {\bf State} & Object state & Open/closed, intact/broken, dry/wet \\
    {\bf Spatial} & Spatial arrangement & Left/right, front/back, occlusion, relative positions \\
    {\bf Saliency} & Visual saliency & Main subject omission, partial missing \\
    {\bf Temporal} & Temporal logic & Order, sequence of steps, occurrence \\
    {\bf Camera} & Camera language & Viewing angle, camera movement, shot type \\
    {\bf Logical} & Logical reasoning & Causality, containment, contrast \\
    {\bf World Knowledge} & Commonsense & Object function, character role, physical laws \\
  \bottomrule
  \end{tabular}
\end{table}

\begin{table}[p]
  \caption{A comprehensive taxonomy of audio-text consistency negative samples, outlining the primary categories, secondary dimensions of mismatch, and detailed descriptions for each type.}
  \label{tab:audiotext_negative_samples}
  \centering
  \scriptsize 
  \renewcommand{\arraystretch}{0.9} 
  \begin{tabular}{@{} p{2.8cm} p{3.5cm} p{5.2cm} @{}}
    \toprule
    Primary Category & Secondary Dimension & Description \\
    \midrule
    {\bf Speech Attributes} 
    & Speaker Identity & Change in speaker identity (gender, age, timbre, accent) \\
    & Speaker Count & Change in number of speakers (monologue $\leftrightarrow$ dialogue $\leftrightarrow$ group) \\
    & Speaking Role & Change in speaking role (narrator, host, reporter, lecturer) \\
    
    \midrule
    {\bf Speech Content} 
    & Semantic Polarity & Semantic polarity reversal (positive $\leftrightarrow$ negative) \\
    & Entity Reference & Entity reference error (person/place/object replacement) \\
    & Causality & Causality mismatch \\
    & Emotional Polarity & Emotion direction reversal \\
    
    \midrule
    {\bf Emotion \& Pragmatics}
    & Emotional Intensity & Change in emotional intensity \\
    & Speech Act & Change in pragmatic function (statement $\leftrightarrow$ command) \\
    
    \midrule
    {\bf Counting \& Degree} 
    & Numerical Value & Numerical change \\
    & Frequency & Frequency change \\
    & Degree & Change in degree/intensity \\
    
    \midrule
    {\bf Sound Effects} 
    & Action-Sound Mapping & Action-sound mismatch \\
    & Sound Source & Incorrect sound source \\
    & Material Cue & Material acoustic property mismatch \\
    & Trigger Existence & Presence/absence of key sound effect \\
    & Visual-Audio Sync & Visual-audio sync error \\
    
    \midrule
    {\bf Action--Sound Align.}
    & Temporal Sync & Action-sound temporal mismatch \\
    & Repetition & Repetition pattern change \\
    & Background Presence & Background sound change \\
    
    \midrule
    {\bf Acoustic Environ.}
    & Reverberation & Reverberation environment mismatch \\
    & Distance Cue & Distance cue mismatch \\
    
    \midrule
    {\bf Music Attributes} 
    & Instrumentation & Instrument replacement \\
    & Genre \& Style & Genre/style mismatch \\
    & Tempo \& Rhythm & Tempo/rhythm change \\
    & Emotional Alignment & Music emotion alignment disruption \\
    
    \midrule
    {\bf Music--Content Rel.}
    & Narrative Support & Music-narrative relation mismatch \\
    
    \midrule
    {\bf Audio Saliency} 
    & Foreground/Back. & Foreground/background reversal \\
    & Focus Consistency & Focus stability disruption \\
    
    \midrule
    {\bf Temporal Structure} 
    & Event Order & Event order shuffled \\
    & Continuity & Continuity disruption \\
    
    \midrule
    {\bf Logical (Audio)} 
    & Part-Whole & Part-whole relation disruption \\
    & Completeness & Completeness disruption \\
    & Physical Plausibility& Acoustic physical plausibility disruption \\
    
    \midrule
    {\bf World Knowledge} 
    & Scene Consistency & Scene-audio conflict \\
    & Role Knowledge & Role knowledge conflict \\
    
    \midrule
    {\bf Cultural / Social} 
    & Cultural Cue & Cultural audio cue mismatch \\
    & Social Norms & Social norm conflict \\
    
    \midrule
    {\bf Language \& Annot.} 
    & Language Label & Language label error \\
    & Content Alignment & Description-content mismatch \\
  \bottomrule
  \end{tabular}
\end{table}
\subsection{Negative Sample Examples}
To provide a clearer understanding of our generation strategies, we present concrete examples of the synthesized negative samples for both video and audio modalities.

\begin{figure}[p]
  \centering
  \includegraphics[width=\linewidth]{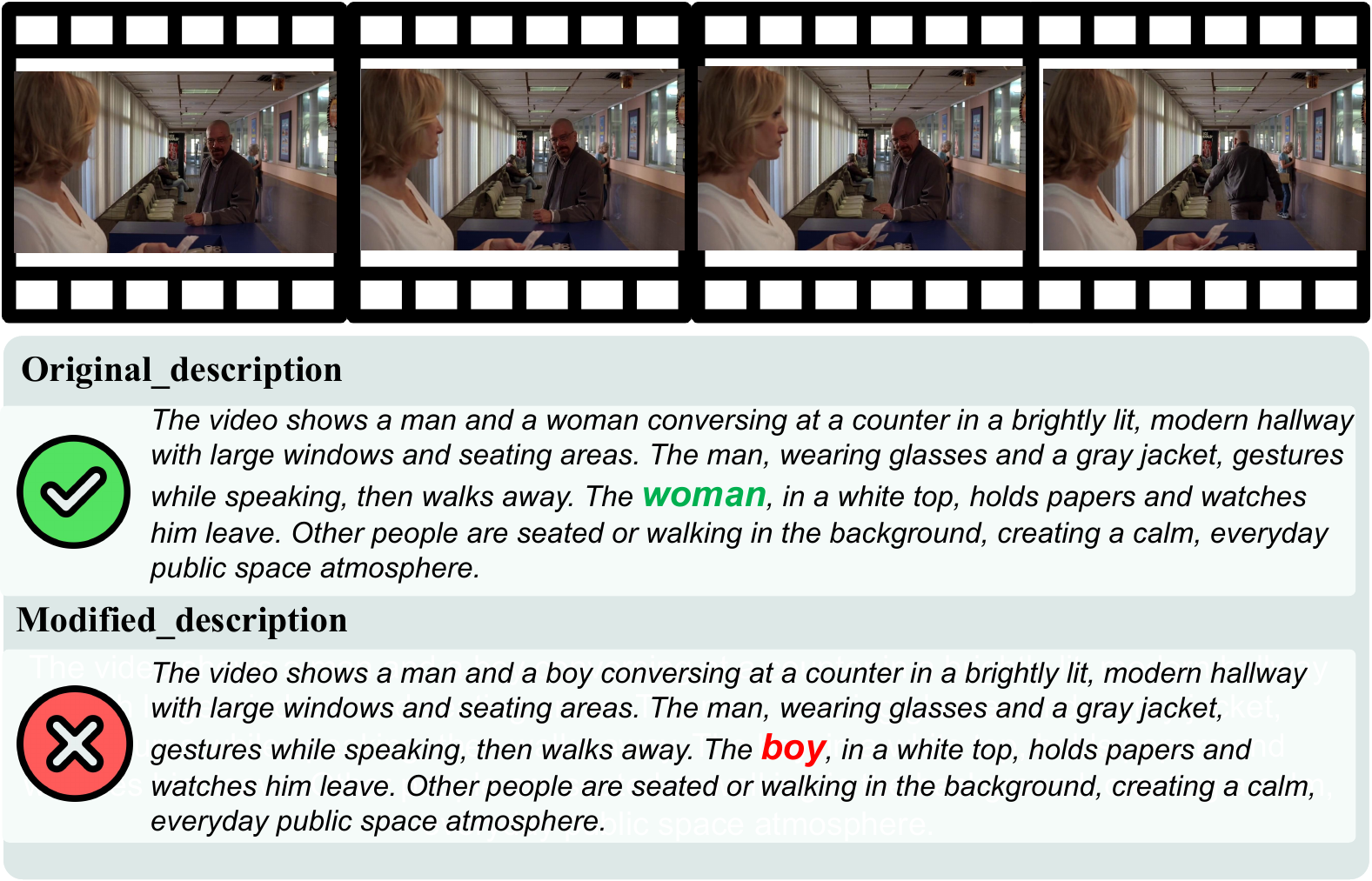}
  \caption{An illustrative example of video-text hard negative construction. The original description accurately depicts a man and a woman. To create a challenging mismatched sample, a minimal semantic error is introduced by replacing the word "woman" with "boy", while keeping the rest of the sentence's grammatical structure perfectly intact.}
  \label{fig:video_text_example}
\end{figure}

Figure \ref{fig:video_text_example} illustrates the construction of a video-text hard negative. In this example, the generation strictly adheres to the syntactic isomorphism constraint by introducing a minimal semantic perturbation—specifically, replacing the word "woman" with "boy". This subtle substitution creates a highly challenging mismatched pair while perfectly preserving the grammatical skeleton and word count of the original description.

\begin{figure}[p]
  \centering
  \includegraphics[width=\linewidth]{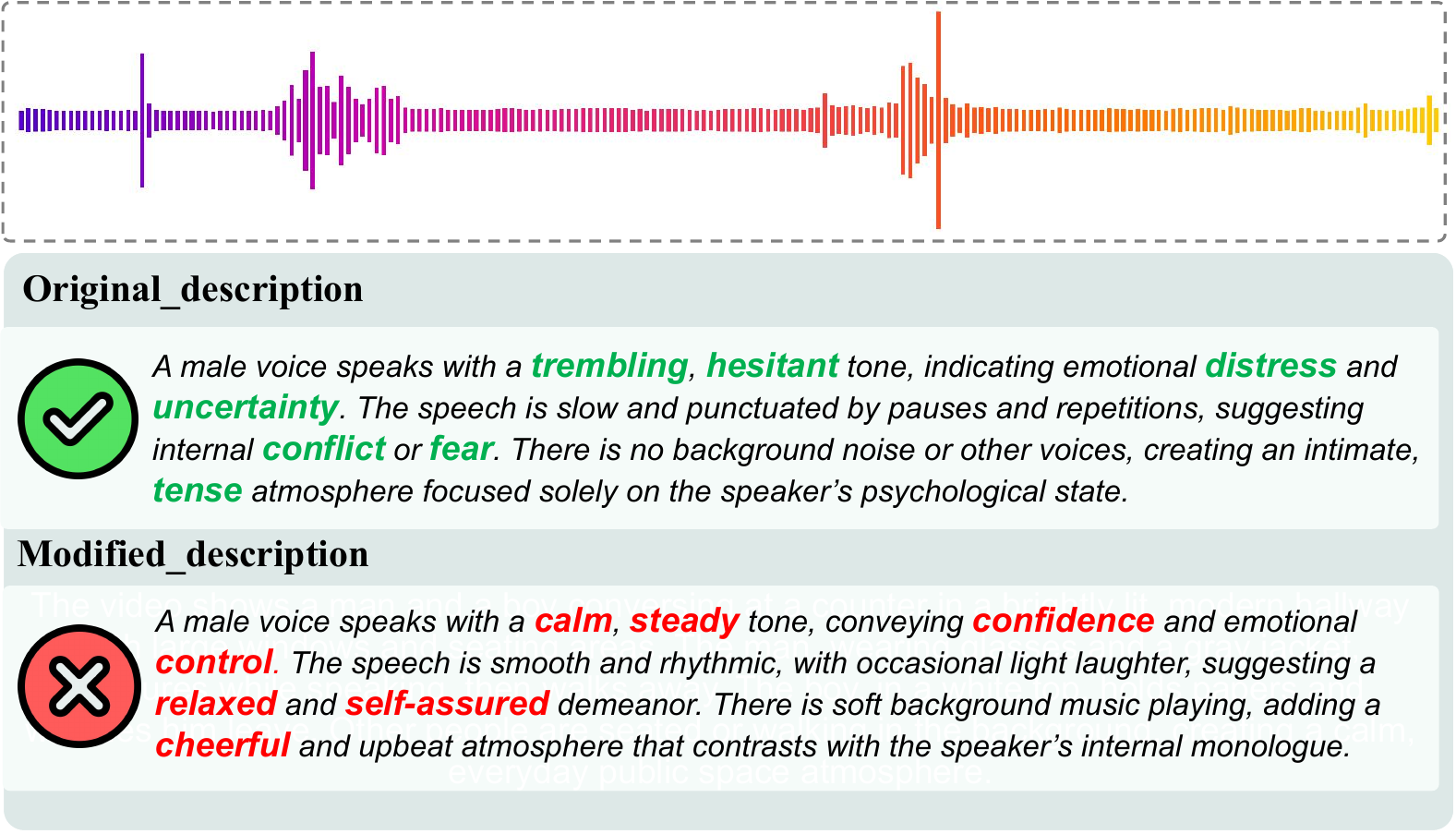}
  \caption{An illustrative example of audio-text mismatched negative construction. The original description depicts a distressed male voice in a silent environment. To create a challenging negative sample, significant semantic and acoustic mismatches are introduced by altering the speaker's emotional state to "calm and steady" and adding "soft background music," creating a clear contradiction while remaining physically plausible.}
  \label{fig:audio_text_example}
\end{figure}

Conversely, Figure \ref{fig:audio_text_example} demonstrates the audio-text negative construction, which relies on more pronounced semantic and acoustic deviations. The original description of a distressed male voice in a silent environment is transformed by altering the emotional state to "calm and steady" and introducing "soft background music." This approach successfully generates a clear, challenging contradiction that completely flips the acoustic narrative while remaining physically and logically plausible.

\section{Extended Details on the Evaluation Dataset}
\begin{figure}[h]
  \centering
  \includegraphics[width=\linewidth]{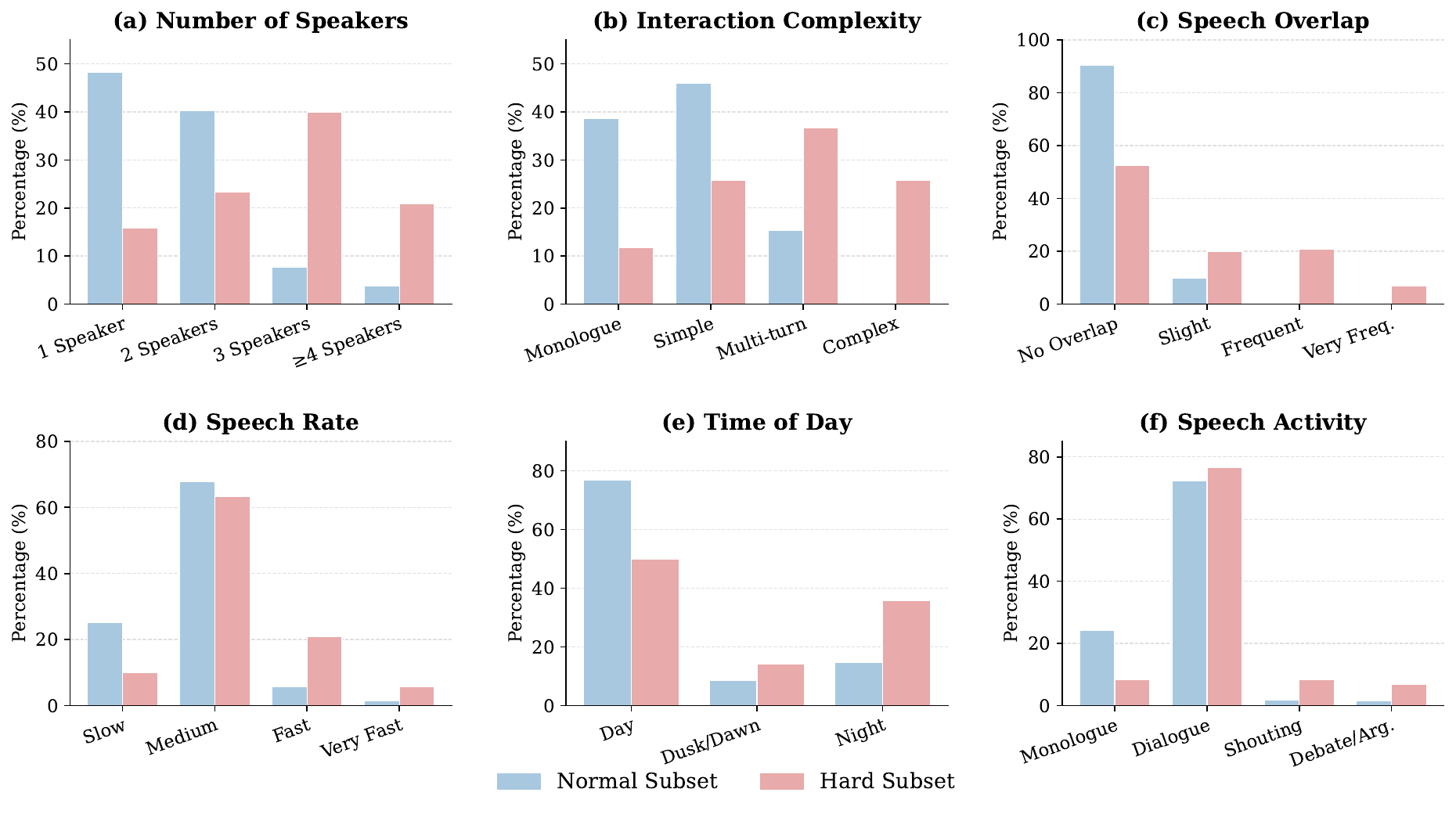}
  \caption{Comparison of data distributions between the \textit{Normal Subset} and \textit{Hard Subset} across: (\emph{a}) Number of Speakers, (\emph{b}) Interaction Complexity, (\emph{c}) Speech Overlap, (\emph{d}) Speech Rate, (\emph{e}) Time of Day, and (\emph{f}) Speech Activity.}
  \label{fig:distribution_shift}
\end{figure}
The evaluation dataset is partitioned into two subsets: a \textbf{Normal Subset} ($N=350$) and a \textbf{Hard Subset} ($N=120$). This structured design is intended to provide a comprehensive assessment of the model's capabilities across varying levels of difficulty. Fig. \ref{fig:distribution_shift} illustrates the significant distribution shifts across six key dimensions.

\subsection{Linguistic and Interaction Complexity}
The dataset exhibits high linguistic diversity, covering 15 languages in the Normal subset and 12 in the Hard subset. 
\begin{itemize}
    \item \textbf{Speaker Density:} A major shift occurs in the number of participants. While the Normal subset is dominated by single-speaker scenarios (48.3\%) , the Hard subset is characterized by multi-speaker environments, with $\geq3$ speakers accounting for 60.8\% of the samples.
    \item \textbf{Interaction Depth:} Interaction complexity moves from simple interactions (46.0\%) in the Normal set to a focus on \textit{Multi-turn Coherent} (36.7\%) and \textit{Complex Multi-person Overlap} (25.8\%) in the Hard set.
    \item \textbf{Speech Overlap:} In the Normal set, 90.3\% of samples contain no overlap. Conversely, the Hard set introduces significant challenges with frequent overlap and interruptions appearing in 47.5\% of the data.
\end{itemize}

\subsection{Acoustic and Visual Environmental Challenges}
The Hard subset is engineered to simulate "in-the-wild" difficulties that stress audio-video alignment:
\begin{itemize}
    \item \textbf{Temporal Dynamics:} The speech rate shifts from predominantly moderate (67.7\%) to include a higher proportion of \textit{Fast} and \textit{Very Fast} speech in the Hard subset (26.6\% total).
    \item \textbf{Lighting and Composition:} Visual difficulty is heightened in the Hard subset through a larger share of \textit{Night} and \textit{Evening} scenes (53.3\% combined) , compared to the Normal set where daytime scenes are more prevalent (42.0\%).
    \item \textbf{Acoustic Background:} The audio environment in the Normal set is mostly \textit{Clean} (45.4\%). The Hard set introduces complex \textit{Crowd Noise} (22.5\%) and \textit{Outdoor Ambience} (18.3\%), demanding higher noise-robustness from the models.
\end{itemize}

\subsection{Character Attributes and Emotional Breadth}
Both subsets maintain a wide range of human attributes. The Hard subset emphasizes intense emotional states, with \textit{Tense} (25.8\%) and \textit{Excited} (14.2\%) being the most frequent non-neutral emotions, whereas the Normal set features a broader distribution of \textit{Serious} and \textit{Happy} states. This emotional diversity, coupled with varying vocal textures like \textit{Sharp} and \textit{Hoarse}, provides a rigorous benchmark for fine-grained audio-video perception.

\section{Additional Experimental Results}
\subsection{Human Preference Prediction Accuracy}
\label{subsec:human_accuracy}

Building upon the model-level Pearson correlation analysis in Section 4.3 of the main paper, we further evaluate the reliability of AVBench by calculating the instance-level prediction accuracy of our automated metrics against human pairwise preferences. While the correlation coefficient validates the overall trend of win ratios, the accuracy metric directly measures how frequently the automated evaluator agrees with human experts in a direct side-by-side comparison (2AFC).

For each pair of videos generated from the same prompt, we define the human consensus as the ground-truth preference. The automated metric's prediction is considered accurate if it assigns a strictly higher score to the video preferred by the human experts. 

\begin{figure}[h]
  \centering
  \includegraphics[width=0.8\linewidth]{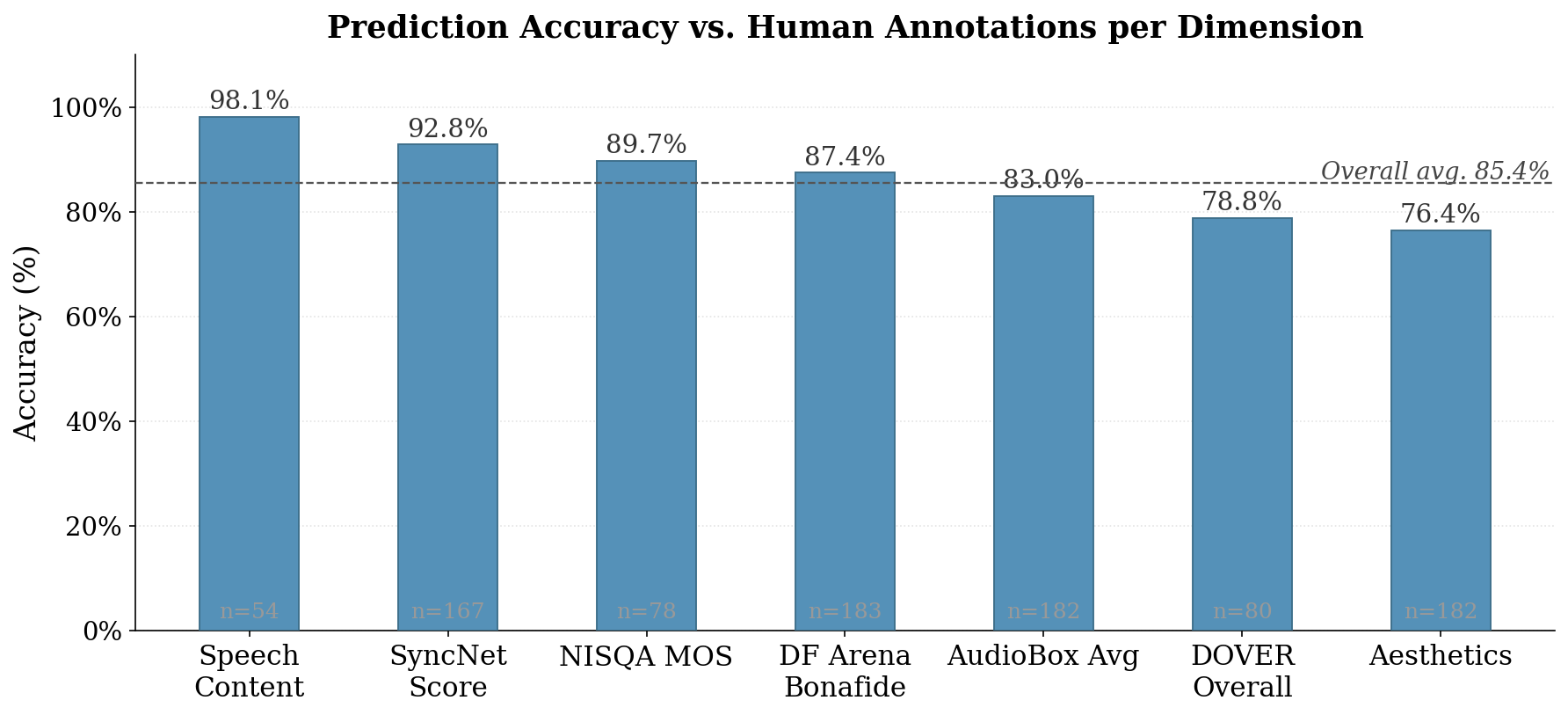}
  \caption{Prediction accuracy of AVBench's automated metrics compared to human expert preferences across seven objective evaluation dimensions. The bar chart displays the percentage of instances where the automated metric correctly assigned a higher score to the human-preferred video in a 2AFC setup (ties excluded). The framework achieves an overall average accuracy of 85.4\%, peaking at 98.1\% for Speech Content, indicating a highly robust capability to mirror human perceptual judgments.}
  \label{fig:accuracy_bar}
\end{figure}

As illustrated in Figure \ref{fig:accuracy_bar}, AVBench consistently achieves high prediction accuracy across seven objective evaluation dimensions. The automated metrics exhibit strong discriminative capabilities, achieving an overall average accuracy of 85.4\% and peaking at 98.1\% for Speech Content. 

\begin{figure}[h]
  \centering
  \includegraphics[width=0.8\linewidth]{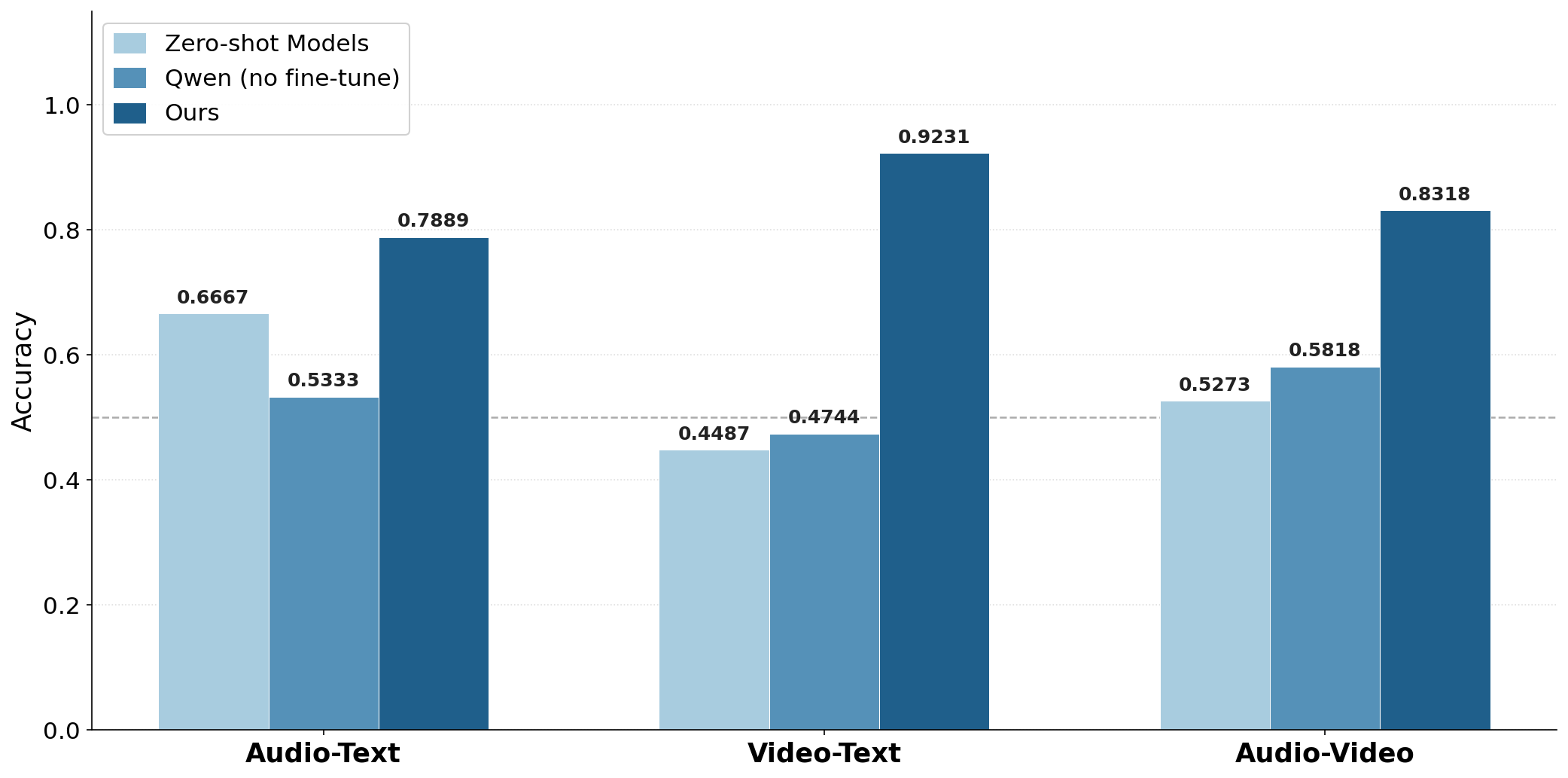}
  \caption{Comparison of prediction accuracy for multi-modal consistency dimensions. The grouped bar chart evaluates the performance of our SFT-trained model ("Ours") against zero-shot models and the base Qwen model across Audio-Text, Video-Text, and Audio-Video consistency. Our fine-tuned framework demonstrates significant improvements over the baselines, peaking at 92.31\% for Video-Text consistency, successfully validating the efficacy of our fine-tuning strategy.}
  \label{fig:consistency_accuracy_bar}
\end{figure}

Furthermore, to evaluate the complex multi-modal alignment capabilities, we compare our fine-tuned evaluator against zero-shot models and the base Qwen model (without fine-tuning) across the three consistency dimensions. As shown in Figure \ref{fig:consistency_accuracy_bar}, our SFT-trained model ("Ours") significantly outperforms the baselines. Notably, in Video-Text consistency, our model achieves an accuracy of 92.31\%, compared to merely 47.44\% for the base model. These instance-level accuracy results demonstrate that our specialized fine-tuning procedure enables the evaluation framework to reliably approximate human judgment, effectively capturing subtle differences in fine-grained consistency.

\subsection{Qualitative Case Studies}
\label{subsec:case_studies}

To further illustrate the effectiveness of our proposed evaluation framework, we present three qualitative case studies covering audio-text, video-text, and audio-video consistency. These examples demonstrate how existing baseline models frequently fail to detect fine-grained mismatches, whereas our SFT-enhanced evaluator consistently aligns with human perception.

\begin{figure}[htb]
  \centering
  \includegraphics[width=\linewidth]{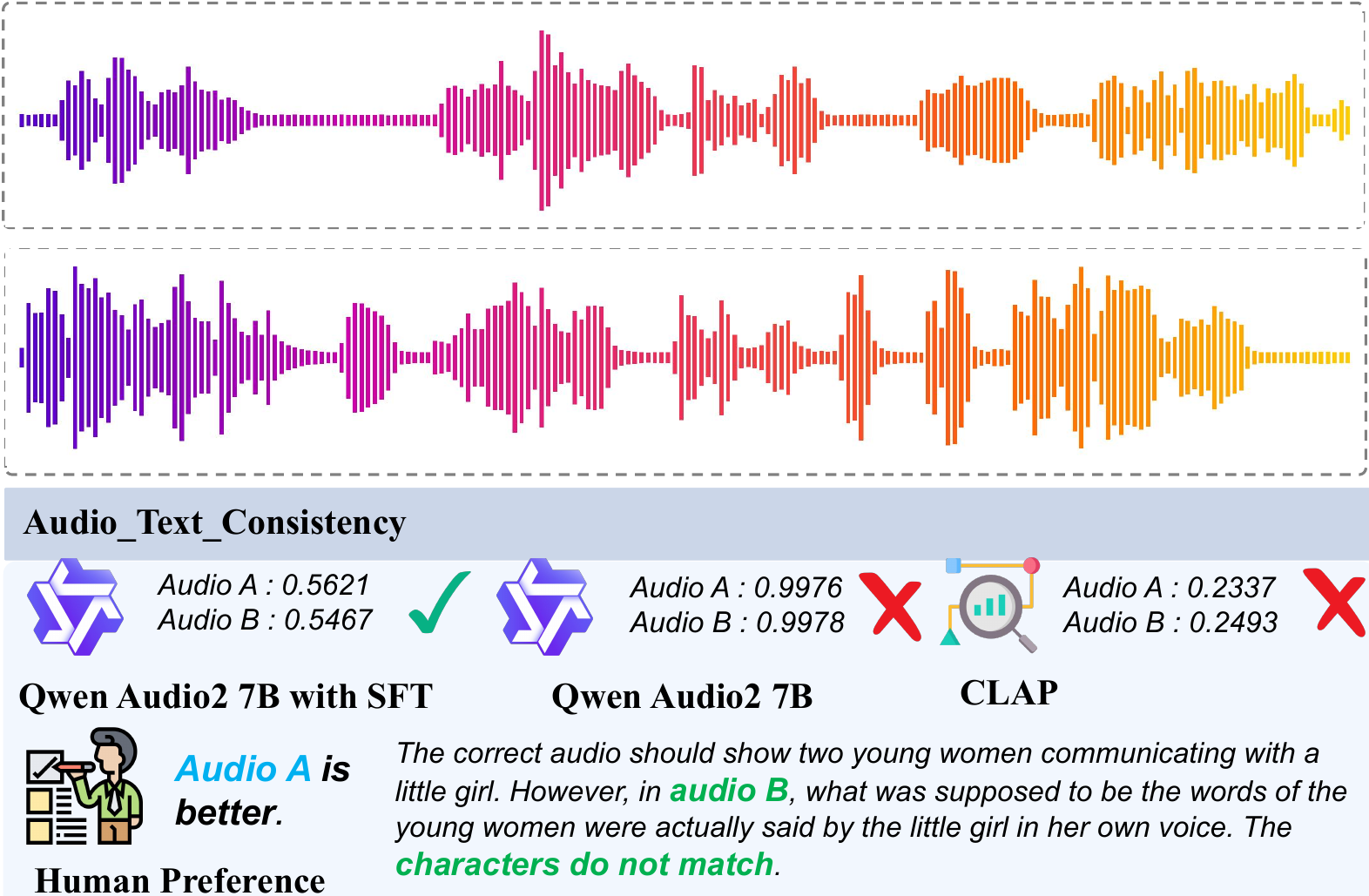}
  \caption{A case study on audio-text consistency evaluation. The original text describes two young women communicating with a little girl. In Audio B, the young women's dialogue is incorrectly spoken by the little girl, creating a character mismatch. While baseline models like CLAP and the base Qwen Audio2 7B fail to recognize this error, our SFT-enhanced evaluator correctly penalizes Audio B and aligns with the human preference for Audio A.}
  \label{fig:audio_text_case}
\end{figure}

Figure \ref{fig:audio_text_case} presents an audio-text consistency scenario involving a speaker identity mismatch. The text describes two young women communicating with a little girl. However, in the negative sample (Audio B), the little girl incorrectly speaks the dialogue meant for the young women. While baseline models such as CLAP and the base Qwen Audio2 7B fail to detect this semantic error and erroneously assign higher scores to Audio B, our SFT-enhanced model correctly identifies the discrepancy, perfectly aligning with the human preference for Audio A.

\begin{figure}[htb]
  \centering
  \includegraphics[width=\linewidth]{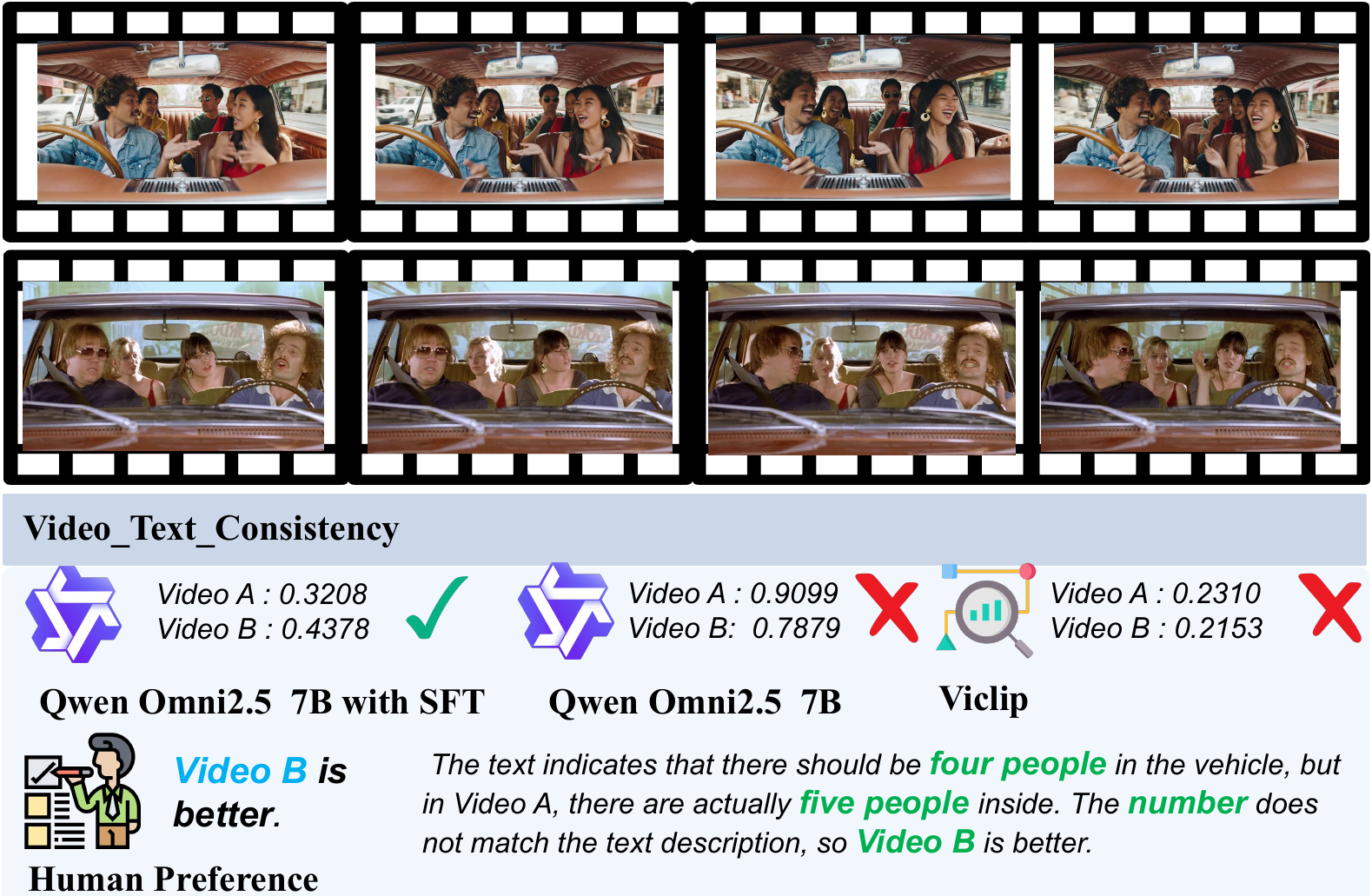}
  \caption{A case study on video-text consistency evaluation, our SFT-enhanced evaluator correctly penalizes Video A and aligns perfectly with the human preference for Video B.}
  \label{fig:video_text_case}
\end{figure}

Similarly, Figure \ref{fig:video_text_case} highlights a video-text evaluation focusing on fine-grained numerical reasoning. The prompt explicitly specifies "four people" in a vehicle, yet Video A clearly depicts five. Baseline models like ViCLIP and the base Qwen Omni2.5 7B overlook this subtle counting error. Conversely, our model successfully captures the numerical mismatch and penalizes the incorrect video, matching human judgment.

\begin{figure}[htb]
  \centering
  \includegraphics[width=\linewidth]{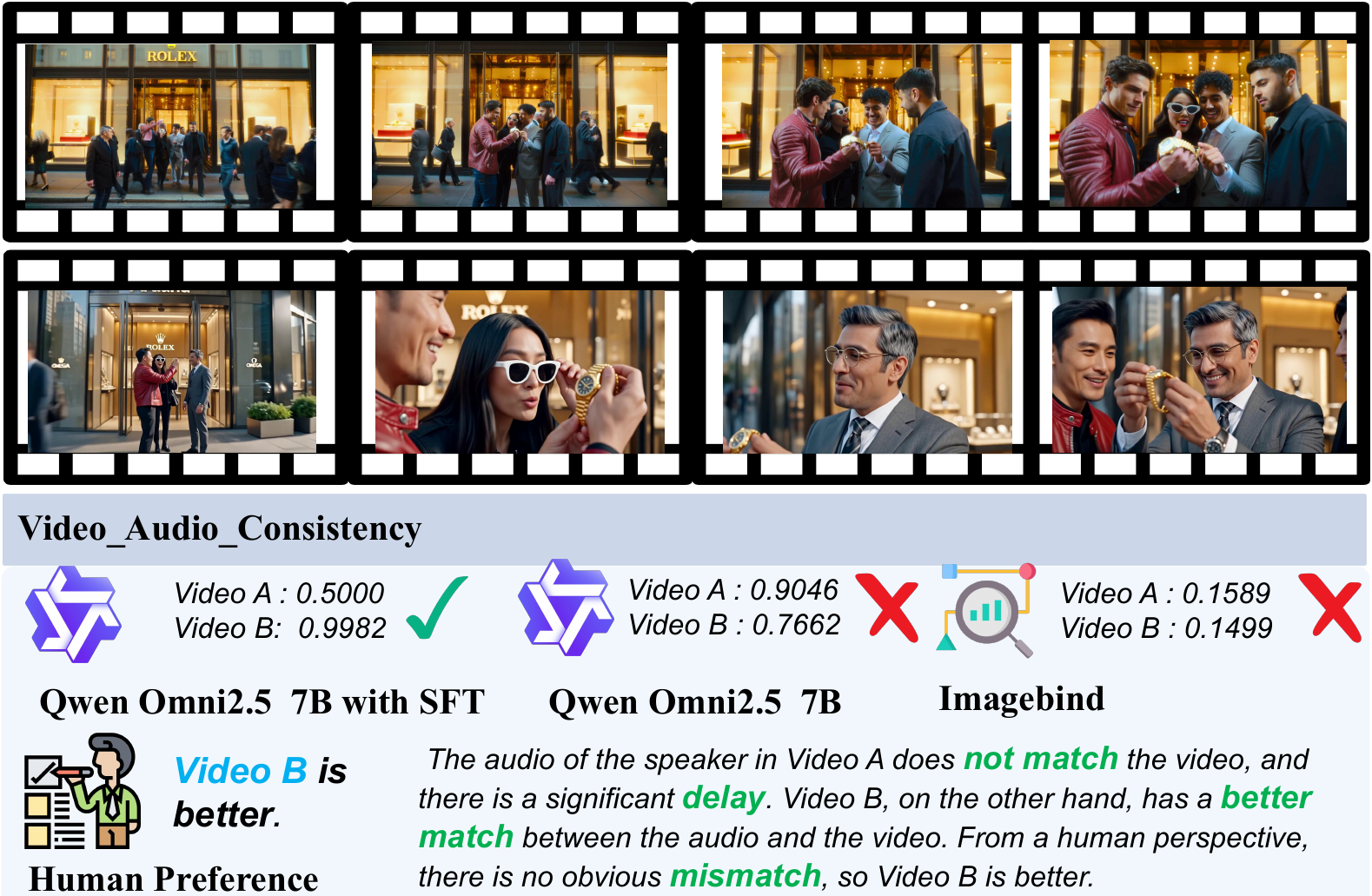}
  \caption{A case study on audio-video consistency evaluation, our SFT-enhanced evaluator successfully identifies the desynchronization and aligns perfectly with human judgment.}
  \label{fig:audio_video_case}
\end{figure}

Finally, Figure \ref{fig:audio_video_case} addresses the temporal aspect of audio-video synchronization. In this example, Video A suffers from a noticeable audio-video delay. Traditional baselines (ImageBind and base Qwen Omni2.5 7B) struggle to capture this temporal misalignment and incorrectly score the delayed video higher. Our SFT-trained evaluator, however, accurately detects the desynchronization and correctly prefers the well-synchronized Video B. Collectively, these cases highlight the robustness of our approach in capturing complex, multi-dimensional consistency errors that traditional metrics miss.


\end{document}